\documentclass[sn-mathphys]{sn-jnl}% Math and Physical Sciences Reference Style
\jyear{2021}%

\theoremstyle{thmstyleone}%
\newtheorem{theorem}{Theorem}%  meant for continuous numbers
\newtheorem{proposition}[theorem]{Proposition}% 

\newtheorem{corollary}{Corollary}

\theoremstyle{thmstyletwo}%
\newtheorem{example}{Example}%
\newtheorem{remark}{Remark}%

\theoremstyle{thmstylethree}%
\newtheorem{definition}{Definition}%

\usepackage{booktabs} % For formal tables
\usepackage{comment}
\usepackage{bm}
\usepackage{epsfig}
\usepackage{color}
\usepackage{graphics}
\usepackage{tabularx}
\usepackage{times}
\usepackage{amsmath}
\usepackage{amsfonts}
\usepackage{algorithm}
\usepackage{here}
\newcommand{\argmax}{\mathop{\rm argmax}\limits}
\newcommand{\argmin}{\mathop{\rm argmin}\limits}
\usepackage{listings}

\begin{document}

\title[Detecting Signs of Model Change]{
Detecting Signs of Model Change with Continuous Model Selection Based on Descriptive Dimensionality}


\author*[1]{\fnm{Kenji} \sur{Yamanishi}}\email{yamanishi@g.ecc.u-tokyo.ac.jp}

\author[1]{\fnm{So} \sur{Hirai}}\email{So.Hirai.16@gmail.com}


\affil*[1]{\orgdiv{School of Information Science and Technology}, \orgname{The University of Tokyo}, \orgaddress{\street{7-3-1 Hongo, Bunkyoku}, \city{Tokyo}, \postcode{113-8656}, \state{}, \country{Japan}}}

%\affil[2]{\orgdiv{Department}, \orgname{Organization}, \orgaddress{\street{Street}, \city{City}, \postcode{10587}, \state{State}, \country{Country}}}

%\affil[3]{\orgdiv{Department}, \orgname{Organization}, \orgaddress{\street{Street}, \city{City}, \postcode{610101}, \state{State}, \country{Country}}}


\abstract{
We address the issue of detecting  changes of  models that lie behind a data stream.
The model refers to an integer-valued structural information such as the number of free parameters in a parametric model. 
Specifically we are concerned with the problem of how we can detect signs of model changes earlier than they are actualized.
To this end, we employ {\em continuous model selection} on the basis of the notion of {\em descriptive dimensionality}~(Ddim). It is a real-valued model dimensionality, which is designed for quantifying the model dimensionality in the model transition period.  {Continuous model selection is to determine the real-valued model dimensionality in terms of Ddim from a given data.}
We propose a novel methodology for detecting signs of model changes by tracking the rise-up of Ddim in a data stream.
We apply this methodology to detecting signs of changes of the number of clusters in a Gaussian mixture model and those of the order in an auto regression model.
With synthetic and real data sets, we empirically demonstrate its effectiveness by showing that it is able to visualize well how rapidly model dimensionality moves in the transition period and to raise early warning signals of model changes earlier than they are  detected with existing methods.}

\keywords{
model change detection, change sign detection, minimum description length principle, model selection, continuous model selection, descriptive dimension, clustering}
%\ \ \\
%{\footnotesize \bf Competing interests:} No.}

\maketitle

\section{Introduction}
\subsection{Motivation}
This paper is concerned with the issue of detecting changes of a model that lies behind a data stream.
{The {\em model} refers to the discrete structural information such as the number of free parameters in the mechanism for generating the data.}
We consider the situation where a model changes over time.
Under this environment, 
it is important to detect the model  changes as accurately as possible. 
This is because the model changes may correspond to important events. 
%They have also been studied in the scenario of {\em dynamic model selection}~(DMS)\cite{ym07, ym05}.
%
For example, it is reported in \cite{hirai} that when customers' behaviors are modeled using a Gaussian mixture model, the change of the number of mixture components %, estimated with  DMS, 
corresponds to the emergence or disappearance of a cluster of customers' behaviors. In this case a model change implies a change of the market trend. 
For another example, it is reported in \cite{ym05} that when the syslog behaviors are modeled using a mixture of hidden Markov models, the change of the number of mixture components may correspond to 
%the emergence or disappearance of a system behavior pattern. In this case, a model change may correspond to 
 a system failure.

The issue of model change detection has extensively been explored.  This paper is rather concerned with the issue of detecting {\em signs} or {\em early warning signals} of model changes. 
Why is it important to detect such signs?
One reason is that if they were detected earlier than the changes themselves, we could predict the changes before they were actualized.
The other reason is that if they were detected after the change themselves, we could analyze the cause of the changes in a retrospective way. 

\begin{figure}[!hbth]
	%\vskip 0.2in
%\centering
%\begin{minipage}%{0.4\hsize}
	\begin{center}
%\centering
		%\centerline{
\includegraphics[keepaspectratio, height=50mm %width=60mm, height=36.0mm
%width=\hsize %
%height=28mm
]{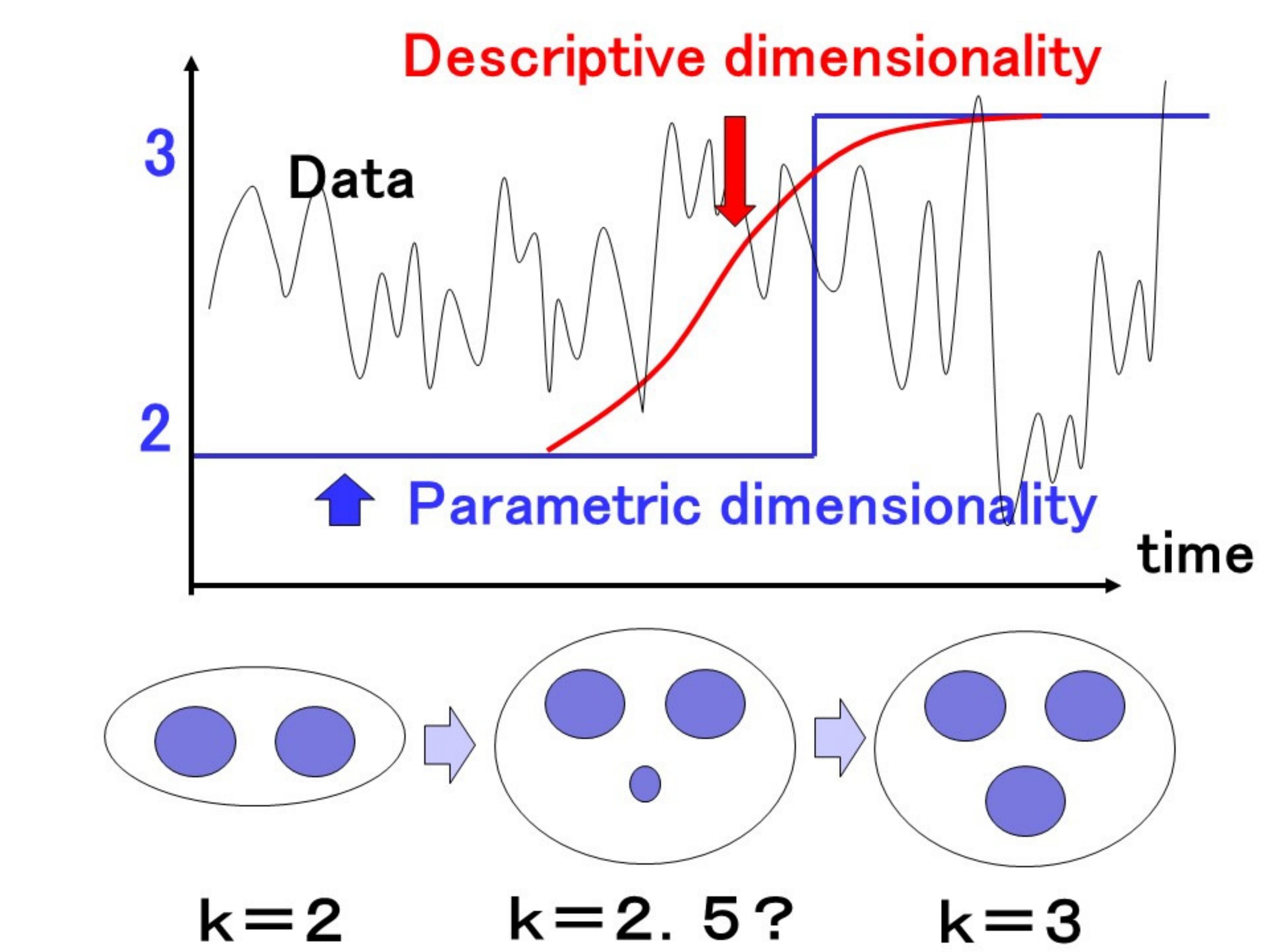} %}% windowsize_auc-eps-converted-to.pdf}}
		\caption{Transition Period of Dimensionality Change}
		\label{ddim}
	\end{center}
%\end{minipage}
\end{figure}

A model, say, the number of parameters, is an integer-valued index, in general. Therefore, it appears that the model change abruptly occurs. However, it is reasonable to suppose that some intrinsic change, which we call {\em latent change},  gradually occurs at the back of the model change. Then we may define a {\em sign} of the model change as the starting point of the latent change.
Therefore, if we properly defined a real-valued index to quantify the model dimensionality in the transition period,  we could understand how rapidly the latent change was going on and we could detect signs of model changes by tracking the rise-up of the index~(Fig. \ref{ddim}). 

The key idea of this paper is to employ the notion of {\em descriptive dimensionality}~(Ddim) %\cite{ddim} 
for the quantification of a model in the transition period.
Ddim is a real-valued index, which quantifies the model dimensionality for the case where a number of models are mixed. 
We thereby establish a methodology of {\em continuous model selection}. { It is to determine the optimal real-valued model dimensionality from data on the basis of Ddim.}
In the transition period of model changes, the mixing structure of models may change over time.
Hence, by tracking the rise-up of Ddim, we will be able to track the latent changes behind model changes. 
%This paper gives the first application of Ddim to model change sign detection.

The purpose of this paper is twofold:
One is to establish a novel methodology for detecting signs (or early warning signals) of model changes from a data stream. We realize this by using Ddim for the quantification of model dimensionality in its transition period. {The theory of Ddim is developed on the basis of the {\em minimum description length}~(MDL) principle~\cite{ris}.
% in combination of the theory of box counting dimension\cite{dudley,man,farmer}.}
%Specifically we consider a Gaussian mixture model as a knowledge representation where a model is the number of components in the mixture and  give details of our methodology applied to GMMs.
The other is to empirically validate the effectiveness of the methodology using synthetic and real data sets. We evaluate how early and how reliably it is able to make alarms of signs of model changes.

\subsection{Related Work}

Model change detection has been studied in the scenario of 
{\em dynamic model selection}~(DMS) developed in \cite{ym05,ym07}.
%Model change detection is different from conventional change detection or 
%in that the former is concerned with changes of latent structural information lying behind data such as the number of parameters, while the latter  is concerned with changes of distributions with regard to some distance measure. 
Model change detection is different from the classical  continuous parameter change detection.
Taking an example of finite mixture models, the former is to detect changes in the number of components, while the latter 
is to detect those in the real-valued parameters of individual components or mixing parameters. 
 In  \cite{ym05, ym07}, they proposed the DMS algorithm which outputs a model sequence of the shortest description length, on the basis of the 
% {\em minimum description length}~(MDL) 
MDL  principle~\cite{ris}.
They demonstrated its effectiveness from 
the empirical and information-theoretic aspects.
%DMS is concerned with when and how the model  changes over time. It was developed by extending the {\em minimum description length}~(MDL) criterion~\cite{rissanen} that was explored in the stationary setting into the non-stationary one.
%DMS was applied to data mining issues such as network failure detection~\cite{ym05}, clustering change detection~\cite{hirai}, %network change detection~\cite{hayashi},
%etc.
%and NMF rank change detection~\cite{ito}.
The MDL based model change detection has been further theoretically  justified in \cite{yamanishi3}.
The problems similar to model change detection have been discussed in the scenarios of switching
distributions~\cite{erven}, %derandomization~\cite{vovk}, 
tracking best experts~\cite{herbster}, on-line clustering~\cite{song},  cluster evolution~\cite{fingerprint}, Bayesian change detection~\cite{xuan}, and structure break detection for autoregression model~\cite{davis}.
% and MDL model change statistics~\cite{yamanishi3}.
In all of these previous studies, however, a model change was considered to be an abrupt change of a discrete structure. 
The transition period  of changes has never been  analyzed.
{In the conventional state-space model, change detection of continuous states is addressed (see e.g.\cite{ding2}). Then the state itself has not the same meaning as a model which we define in this paper. The number of states is a model which we mean.}

Changes that do not occur abruptly but
 incrementally occur were discussed in the context of detecting incremental changes in concept drift~\cite{gama}, %continuous change~\cite{miyaguchi}, 
 gradual changes~\cite{yamanishi2}, volatility shift~\cite{volatility}, etc.
However, it has never been quantitatively analyzed how rapidly a model changes in the transition period.

Recently, the indices of structural entropy~\cite{bigdata2018} and graph-based entropy~\cite{ohsawa} have been developed for measuring the uncertainty associated with model changes. 
%The MDL change statistics has been proposed in 
%
 Although they can be thought of as early warning signals of model changes, they cannot explain the intrinsic model dimensionality nor how rapidly a model changes in the transition period. 
Change sign detection method usng differential MDL change statistics has been proposed in \cite{sr}. However, it is applied to change sign detection for parameters only not for models.

This paper proposes a methodology for analyzing model transition in terms of real-valued dimensionality.
A number of notions of dimensionality have been proposed
in the areas of physics and statistics.
The metric dimension was proposed by Kolmogorov and Tihomirov~\cite{kol} to measure the complexity of a given set of points in terms of the notion of covering numbers. This was evolved into the notion of  the box counting dimension, equivalently, the fractal dimension~\cite{man}. 
It is a real-valued index for quantifying the complexity of  a given set. 
It is also related to the capacity~\cite{dudley}. 
Vapnik Chervonenkis dimension was proposed to measure the power of representation for a given class of functions~\cite{vapnik}.
It was also related to the rate of uniform convergence of estimating functions. See \cite{Haussler} for relations between dimensionality and learning.
{The dimensionality as a power of representation is conventionally integer-valued, but when it changes over time, there is no effective non-integer valued quantification of its transition.}
%However, any notion appropriate for quantifying the model dimensionality in the model transition period has never existed.

\begin{comment}
Recently,  a theory of {\em descriptive dimensionality} (Ddim) has been developed from a view of information theory~\cite{ddim}. Ddim is a real-valued index that measures the model complexity for the case where a number of different models are mixed. Ddim coincides with the conventional dimensionality (the number of free parameters) for the case where a model is a single parametric one.
Ddim has turned out to be the intrinsic dimensionality that characterizes the performance of the MDL-based learning and change detection. 
Although Ddim is expected to be effective for visualizing the model transition period, it has never been applied to visualizing transition periods of model changes or raising their early warning signals.
\end{comment}

\subsection{Significance of This Paper}

The significance of this paper is summarized as follows:\\
(1){\em Proposal of a novel methodology for detecting signs of model changes with continuous model selection.}
This paper proposes a novel methodology for detecting signs of model changes. 
The key idea is to track model transitions with continuous model selection using the notion of {\em descriptive dimensionality} ~(Ddim). 
It measures the model dimensionality in the case where a number of models with different dimensionalities are mixed.

For example, we employ the Gaussian mixture model (GMM) to consider the situation where the number of mixture components changes over time. 
We suppose that in the transition period of model change,  
a number of probabilistic models with various mixture sizes are fused.
We give a method for calculating Ddim for this case.
The transition period of model change 
can be visualized by drawing a Ddim graph versus time.
Once a Ddim graph is obtained, we can understand how rapidly the model changes over time. 
We eventually detect signs of model changes by tracking the rise-up of Ddim. 
This methodology is significantly important in data mining since it helps us predict model changes  in earlier stages.

(2){\em Empirical demonstration of effectiveness of model change sign detection via Ddim.} 
We empirically validate how early we are able to detect signs of model changes with continuous model selection, for GMMs and auto-regression~(AR) models. 
With synthetic data sets and real data sets, 
we illustrate that our method is able to effectively visualize the transition period of model change using Ddim.
We further empirically demonstrate that our methodology is able to 
detect signs of model changes
significantly  earlier than any existing dynamic model selection algorithms and is comparable to structural entropy in \cite{bigdata2018}. 
Through our empirical analysis, we demonstrate that Ddim is an effective index for measuring the model dimensionality in the model transition period.

(3){\em Giving theoretical foundations for Ddim.} 
In this paper, Ddim plays a central role in continuous model selection. We introduce this notion from an information-theoretic 
view based on the MDL principle~\cite{ris} (see also \cite{grunwald}). 
We show that Ddim coincides with the number of free parameters in the case where the model consists of a single parametric class. 
We also derive Ddim for the case where a number of models with different dimensionalities are mixed.
We characterize Ddim by demonstrating that it governs the rate of convergence of  the MDL-based learning algorithm. 
This corresponds to the fact that the metric dimensionality governs the rate of convergence of the empirical risk minimization algorithm in statistical learning theory~\cite{Haussler}.

A preliminary version of this paper appeared in Arxiv \cite{ddim},  in which only the theoretical foundation of Ddim was given and no contents on continuous model selection with its applications to model change sign detection was included. 
This paper is  a complete version of \cite{ddim} including not only the theory of Ddim but also its applications.

The rest of this paper is organized as follows: Sec. 2 introduces the notion of Ddim. Sec. 3 gives a methodology for model change sign detection via Ddim. Sec. 4 shows experimental results. 
Sec. 5 characterizes Ddim by relating it to the rate of convergence of the MDL learning algorithm. Sec. 6 gives conclusion.
Source codes and data sets are available at a Github repository~\cite{dit}.  
%Supplementary materials show how to use the program.
%%%%%%%%%%%%%%%%%%%%%%%%%%%%%%%%%%%%%%%%%%
\section{Descriptive Dimensionality}

\subsection{NML and Parametric Complexity}

This section introduces the theory of Ddim. % according to \cite{ddim}. 
This theory is based on the MDL %~(Minimum Description Length) 
principle (see \cite{ris} for the original paper and \cite{rissanen} for the recent advances) from the viewpoint of information theory.
We start by introducing a number of fundamental notions of the MDL principle.

Let ${\mathcal X}$ be the data domain where ${\mathcal X}$ is either discrete or continuous. Without loss of generality, we assume that ${\mathcal X}$ is discrete.
Let 
${\bm x}=x_{1},\dots ,x_{n}\in {\mathcal X}^{n}$ be a data sequence of length $n$.  We assume that each $x_{i}$ is independently generated. 
%Hereafter, we write $x^{n}$ as ${\bm x}$ for the sake of notational simplicity.
${\mathcal P}=\{p({\bm x}) \}$ be a class of probabilistic models where $p({\bm x})$ is a probability mass function or a probability density function.

Under the MDL principle, the information of a datum ${\bm x}$ is measured in terms of description length, i.e, the codelength required for encoding the datum with a prefix coding method.
We may encode ${\bm x}$ with help of a class ${\mathcal P}$ of probability distributions.
One of the most important methods for calculating the codelength ${\bm x}$ using ${\mathcal P}$ is the normalized maximum likelihood~(NML) coding~\cite{rissanen}.
This is defined as the codelength associated with the NML distribution as:

\begin{definition}{\rm 
We define the {\em normalized maximum likelihood (NML) distribution} over ${\mathcal X}^{n}$ with respect to ${\mathcal P}$ by
\begin{eqnarray}\label{nmld}
p_{_{\rm NML}}({\bm x};{\mathcal P})\buildrel \rm def \over =\frac{\max _{p\in {\mathcal P}}p({\bm x})}{\sum _{{\bm y}} \max _{p\in {\mathcal P}}p({\bm y})}.
\end{eqnarray}
The {\em normalized maximum likelihood (NML) codelength} of ${\bm x}$ relative to ${\mathcal P}$, which we denote as $L_{_{\rm NML}}({\bm x}; {\mathcal P})$, is given as follows:
\begin{eqnarray}\label{sc0}
L_{_{\rm NML}}({\bm x};{\mathcal P})&\buildrel \rm def \over =&-\log p_{_{\rm NML}}({\bm x}; {\mathcal P})
\nonumber \\
           &=&-\log \max _{p\in {\mathcal P}}p({\bm x})+\log 
           {\mathcal C}_{n}({\mathcal P}),
\end{eqnarray}
where 
\begin{eqnarray}\label{scc}
\log {\mathcal C}_{n}({\mathcal P})\buildrel \rm def \over =\log 
\sum_{{\bm y}} \max _{p\in {\mathcal P}}p({\bm y}).
\end{eqnarray}
}
\end{definition}

The first term in (\ref{sc0}) is the negative logarithm of  maximum likelihood while the second term (\ref{scc}) is the logarithm of the normalization term. 
The latter is called
the {\em parametric complexity} of ${\mathcal P}$~\cite{rissanen}. 
This means the information-theoretic complexity for the model class ${\mathcal P}$ relative to the length $n$ of data sequence.
The NML codelength can be thought of as an extension of Shannon information $-\log p({\bm x})$ into the case where the true model $p$ is unknown but only ${\mathcal P}$ is known. 
%The MDL principle asserts that the model minimizing the NML codelength is the best one to be selected from a given data.

In order to understand the meaning of the NML codelength and the parametric complexity,  
we define the {\em minimax regret} as follows:
\begin{eqnarray*}\label{minimaxregret}
R_{n}({\mathcal P})\buildrel \rm def \over =\min _{q} \max_{{\bm x}}\left\{ -\log q({\bm x})-\min _{p\in {\mathcal P}}(-\log p({\bm x}))\right\},
\end{eqnarray*}
where the minimum  is taken over the set of all probability distributions. 
The minimax regret means the descriptive complexity of the model class, indicating how largely any codelength is 
deviated from the smallest negative
log-likelihood over the model class. 
Shtarkov \cite{shtarkov} proved that the NML distribution (\ref{nmld}) is optimal in the sense that it attains the minimum of the minimax regret. In this sense the NML codelength is the optimal codelength for encoding ${\bm x}$ for given ${\mathcal P}$.
Then we can immediately see that the minimax regret coincides with the parametric complexity. That is,
\begin{eqnarray}\label{equiv}
R_{n}({\mathcal P})=C_{n}({\mathcal P}).
\end{eqnarray}

We next consider how to calculate the parametric complexity.
According to \cite{rissanen} (pp:43-44), the parametric complexity can be rewritten using a variable transformation technique as follows:
\begin{eqnarray}\label{int1}
 C_{n}({\mathcal P})=
 \sum _{{\bm y}}  \max _{p\in {\mathcal P}}p({\bm y})
         =\int g(\hat{p}, \hat{p})d\hat{p},
\end{eqnarray}
where $g(\hat{p},p)$ is defined as 
\begin{eqnarray}\label{gfunc}
g(\hat{p}, p )\buildrel \rm def \over =
\sum
 _{{\bm y}:\max _{\bar{p}\in {\mathcal P}}\bar{p}({\bm y})=\hat{p}({\bm y})}
%_{{\bm y}: \hat{p}=\argmax _{\bar{p}\in {\mathcal P}}\bar{p}({\bm y})}
 p({\bm y}).
\end{eqnarray}

\subsection{Definition of Descriptive Dimensionality}
Below we give the definition of Ddim from a view of approximation of the parametric complexity, equivalently, the minimax regret (by (\ref{equiv})).
The scenario of defining Ddim is as follows:
We first count how many points are required to approximate the parametric complexity (\ref{int1}) with quantization. 
We consider that count as information-theoretic richness of representation for a  model class.
We then employ that count to define Ddim in a similar manner with the box counting dimension.

We consider to approximate (\ref{int1}) with a finite sum of partial integrals of $g(\hat{p},\hat{p})$.
Let 
$\overline{{\mathcal P}}=\{p _{1}, p _{2},\dots\}
\subset {\mathcal P}$ 
be a finite subset of %quantized 
${\mathcal P}$. 
For $\epsilon >0, $ for $p_{i}\in \overline{{\mathcal P}}$, let
$D_{\epsilon}^{n}(i)\buildrel \rm def \over =\{p\in {\mathcal P} :\ d_{n}(p_{i},p)\leq \epsilon ^{2}\}$ where 
$d_{n}(p_{i}, p)$ is the Kullback-Leibler (KL) divergence between $p$ and $p_{i}$: 
\[d_{n}(p, p_{i})=\frac{1}{n}\sum _{{\bm x}}
p_{i}({\bm x})\log \frac{p_{i}({\bm x})}{p({\bm x})}.\]
%\lim _{n\rightarrow \infty}\frac{1}{n}\int %\sum _{x}
%p(x^{n})\log \frac{p(x^{n})}{p_{i}(x^{n})}dx^{n}.\]
 Then we approximate ${\mathcal C}_{n}({\mathcal P})$ by
\begin{eqnarray}\label{approx}
\overline{{C}_{n}}(\overline{{\mathcal P}})\buildrel \rm def \over = \sum_{i=1}^{m_{n}(\epsilon: {\mathcal P})}
Q_{\epsilon}(i),
\end{eqnarray}
where 
\begin{eqnarray}\label{repp}
Q_{\epsilon}(i)\buildrel \rm def \over =\int _{\hat{p}\in D_{\epsilon}^{n}(i)}g(\hat{p}, \hat{p})d\hat{p}.
\end{eqnarray}
That is, (\ref{approx}) gives an approximation to $C_{n}({\mathcal P})$ with a finite sum of integrals of $g(\hat{p}, \hat{p})$ over the 
$\epsilon ^{2}-$neighborhood of a  point $p_{i}$.
% with respect to the KL-divergence.
We define  $m_{n}(\epsilon :{\mathcal P})$ as the smallest number of  points $\mid \overline{\mathcal P}\mid $ with respect to $\overline{\mathcal P}$ such that
$C_{n}({\mathcal P}) \leq \overline{C}_{n}(\overline{{\mathcal P}})$. More precisely,
\begin{eqnarray}\label{approx2}
m_{n}(\epsilon :{\mathcal P})\buildrel \rm def \over =\min _{\overline{{\mathcal P}}}%: C_{n}({\mathcal P})=\overline{C_{n}}(\overline{{\mathcal P}})e^{o(1)}}
\mid \overline{{\mathcal P}}\mid \ \ {\rm subject\ to}\ 
C_{n}({\mathcal P})\leq \overline{C_{n}}(\overline{{\mathcal P}}).
\end{eqnarray}
%where the minimum is taken w.r.t. $\overline{{\mathcal P}}$ s.t. $C_{n}({\mathcal P})=\overline{C_{n}}(\overline{{\mathcal P}})e^{o(1)}$ and $\lim _{n\rightarrow \infty}o(1)=0$.
%\begin{eqnarray}\label{approx2}
%
% \sum _{i=1}^{m_{n}(\epsilon)}Q_{\epsilon}(i).
%\end{eqnarray}
%That is, 
%\begin{eqnarray}\label{approx2}
%m_{n}(\epsilon : {\mathcal P})=\inf _{\bar{\mathcal P}: C_{n}({\mathcal P})\leq \bar{C}_{n}(\bar{\mathcal P}) }|\bar{\mathcal P}|.
%\end{eqnarray}
%$J(\theta )\buildrel \rm def \over =E\left[ -\log p(X;\theta ,k)/\partial \theta \partial \theta ^{T}\right]$: Fisher information matrix\\

We are now led to the definition of descriptive dimension. 
\begin{definition} {\rm \cite{ddim} 
Let ${\mathcal P}$ be a class of probability distributions.
%$X=\{x_{1},\dots , x_{n}\}$\\
We let $m(\epsilon :{\mathcal P})$ be the one obtained by 
 choosing $\epsilon ^{2}n=O(1)$ in $m_{n}(\epsilon :{\mathcal P} )$ as in (\ref{approx2}).
We define the {\em descriptive dimension}~(Ddim) of ${\mathcal P}$
by 
\begin{eqnarray}\label{defdim}
{\rm Ddim}({\mathcal P})\buildrel \rm def \over =\lim _{\epsilon \rightarrow 0}\frac{\log m(\epsilon : {\mathcal P})}{\log (1/\epsilon )}, 
\end{eqnarray}
when the limit exists.
}
\end{definition}

The definition of Ddim is similar with that of the {\em box counting dimension}~\cite{dudley,man,farmer} .
The main difference between them is how to count the number of points.
 Ddim is calculated on the basis of the number of points required for approximating the parametric complexity, 
while the box counting dimension is calculated on the basis of the number of points required for covering a given object with their $\epsilon $-neighborhoods.

%Denoting $m_{n}({\mathcal P})$ as the total number of  representative points for parametric complexity for ${\mathcal P}$ obtained by choosing $\epsilon ^{2}n=O(1)$ in $m(\epsilon :{\mathcal P})$, 
%Eq.(\ref{defdim}) is equivalent with
%\begin{eqnarray}\label{defdim2}
%{\rm Ddim}({\mathcal P})=\lim _{n \rightarrow \infty}\frac{2\log m_{n}({\mathcal P})}{\log  n}.
%\end{eqnarray}

Consider the case where ${\mathcal P}_{k}$ is a $k$-dimensional parametric class, i.e.,
${\mathcal P}_{k}=\{p({\bm x};\theta ):\ \theta \in \Theta _{k}\subset {\mathbb R}^{k}\}$,
where $\Theta _{k}$ is a $k$-dimensional real-valued parameter space.
Let $p({\bm x};\theta )=f({\bm x}\mid\hat{\theta}({\bm x}))g(\hat{\theta}({\bm x});\theta )$ for the conditional probabilistic mass function $f({\bm x}\mid \hat{\theta}({\bm x}))$.
We then write  $g$ according to (\ref{gfunc})  as follows 
\begin{eqnarray}\label{gfunc2}
g(\hat{\theta}, \theta )=\sum _{{\bm x}:\argmax_{\theta}p({\bm x};\theta)=\hat{\theta}}p({\bm x};\theta).
\end{eqnarray}
Assume that the central limit theorem holds for the maximum likelihood estimator of a parameter vector $\theta$.
Then according to \cite{rissanen}, we can
 take a Gaussian density function as (\ref{gfunc2}) asymptotically.
{That is, for sufficiently large $n$, (\ref{gfunc2}) can be approximated as:
\begin{eqnarray}\label{clt}
g(\hat{\theta}, \theta )\simeq \left(\frac{n}{2\pi }\right)^{\frac{k}{2}}\mid I_{n}(\theta)\mid^{\frac{1}{2}}e^{-n(\hat{\theta}-\theta)^{\top}I_{n}(\theta )(\hat{\theta}-\theta )/2},
\end{eqnarray}}
where $I_{n}(\theta )\buildrel \rm def \over =(1/n)E_{\theta}[-\partial ^{2}\log p({\bm x};\theta )/\partial \theta \partial \theta ^{\top}]$ is the Fisher information matrix.

%Under the assumption of (\ref{clt}),
The following theorem shows the basic property of $m_{n}(\epsilon :{\mathcal P}_{k})$ for the parametric case.
\begin{theorem}\label{basic}
Suppose that $p({\bm x};\theta )\in {\mathcal P}_{k}$ is continuously three-times differentiable with respect to $\theta$.
Under the assumption of the central limit theorem so that (\ref{clt}) holds,
for sufficiently large $n$, we have
\begin{eqnarray}\label{nmb1}
\log C_{n}({\mathcal P}_{k}) = \log m_{n}(1/\sqrt{n} :{\mathcal P}_{k})+O(1). %\frac{k}{2}\log (\epsilon ^{2}n)+O(1).  
%-\frac{k}{2}\log \left( \frac{2\epsilon ^{2}n}{k\pi}\right)+
\end{eqnarray}
%Setting $\epsilon $ such that $\epsilon ^{2}n=k$ yields
%\begin{eqnarray}\label{nmb2}
%\log m_{n}(\epsilon) = \log C_{n}({\mathcal P})-\frac{k}{2}\log \left( \frac{2}{\pi}\right).
%\end{eqnarray}
\end{theorem}
\begin{comment}
Then the following theorem shows the basic property of $m_{n}(\epsilon :{\mathcal P})$.
\begin{theorem}\label{basic}%{\cite{ddim}}
Suppose that ${\mathcal P}_{k}$ is a $k$-dimensional parametric class, i.e., 
${\mathcal P}=\{p({\bm x};\theta ):\ \theta \in \Theta _{k}\subset {\bf R}^{k}\}$, 
where $\Theta _{k}$ is a $k$-dimensional real-valued parameter space. Under the condition that the central limit theorem holds for the maximum likelihood estimator of a parameter vector $\theta$,  for sufficiently large $n$, we have
\begin{eqnarray}\label{nmb1}
\log m(\epsilon :{\mathcal P}_{k}) = \log C_{n}({\mathcal P}_{k})-\frac{k}{2}\log \left( \frac{2\epsilon ^{2}n}{k\pi}\right)+%\frac{\epsilon ^{2}n}{2}+
o(1).
\end{eqnarray}
\end{theorem}
\end{comment}
The proof is given in Appendix.

It is known \cite{rissanen} (p.53) that
 under some regularity condition that the central limit theorem holds for the maximum likelihood estimator for $\theta$, the parametric complexity for ${\mathcal P}_{k}$ is asymptotically 
expanded as 
\begin{eqnarray}\label{asymp}
\log C_{n}({\mathcal P}_{k})=\frac{k}{2}\log \frac{n}{2\pi}+\log \int \sqrt{\mid I(\theta )\mid}d\theta +o(1),
\end{eqnarray}
where $I(\theta )$ is the Fisher information matrix:
$I(\theta )\buildrel \rm def \over =\lim _{n\rightarrow \infty}(1/n)\times$ ${\rm E}_{\theta}[-\partial ^{2}\log p({\bm x};\theta )/\partial \theta \partial \theta ^{\top}]$. 
Plugging (\ref{nmb1}) with (\ref{asymp}) for $\epsilon ^{2}n=O(1)$ into (\ref{defdim}) 
yields the following theorem.

\begin{theorem}\label{th1}%{\rm \cite{ddim}} %(Ddim for parametric classes) 
For a $k$-dimensional parametric class ${\mathcal P}_{k}$,
%=\{p({\bm x};\theta ,k):\ \theta \in \Theta _{k}\subset {\bm R}^{k}\}$,  
under the regularity condition for ${\mathcal P}_{k}$ as in Theorem \ref{basic}, % so that the central limit theorem holds for the maximum likelihood estimator of a parameter, 
we have
\begin{eqnarray}
{\rm Ddim}({\mathcal P}_{k})=k.
\end{eqnarray}
\end{theorem}

Theorem \ref{th1} shows that  when the model class is a single parametric one, Ddim coincides with the conventional notion of dimensionality (the number of free parameters), which we call the {\em parametric dimensionality} in the rest of this paper.

 Ddim can also be defined even for the case where the model class is not a single parametric class.
Hence Theorem \ref{th1} implies that Ddim is a natural extension of the parametric dimensionality.

Let us  consider {\em model fusion} where a number of model classes are probabilistically mixed.
Let ${\mathcal F}=\{ {\mathcal P}_{1},\dots , {\mathcal P}_{s}\}$ be a family of model classes and assume a model class is probabilistically distributed according to 
 $p({\mathcal P})$ over ${\mathcal F}$.  
We denote the model fusion over ${\mathcal F}$ as ${\mathcal F}^{\odot }={\mathcal P}_{1}\odot \cdots \odot {\mathcal P}_{s}$. 
{We may interpret the resulting distribution over ${\mathcal X}$ as a finite mixture model~\cite{mp} of a number of model classes with different dimensionalities.}
Then Ddim of ${\mathcal F}^{\odot}$ 
is calculated as%may be defined as
 \begin{eqnarray}\label{fusiondef}
%{\rm Ddim} ({\mathcal F}^{\odot})
%\buildrel \rm def \over =
\lim _{\epsilon \rightarrow 0}\frac{\log E_{{\mathcal P}}[m(\epsilon : {\mathcal P})]}{\log (1/\epsilon )}
& \geq &\lim _{\epsilon \rightarrow 0}\sum ^{s}_{i=1}p({\mathcal P}_{i})\frac{\log m(\epsilon : {\mathcal P}_{i})}{\log (1/\epsilon )}\nonumber \\
&=&
\sum_{i=1}^{s}p({\mathcal P}_{i}){\rm Ddim}({\mathcal P}_{i}),
 \end{eqnarray}
\begin{comment}
We immediately obtain the following lower bound on Ddim of model fusion.
\begin{theorem}\label{fusionddim}
\begin{eqnarray}\label{fusiondef}
{\rm Ddim} ({\mathcal F}^{\odot})\geq \sum_{i=1}^{s}p({\mathcal P}_{i}){\rm Ddim}({\mathcal P}_{i}).
\end{eqnarray}
\end{theorem}
{\em Proof.} We employ the Jensen's inequality to obtain
\begin{eqnarray*}
{\rm Ddim} ({\mathcal F}^{\odot})&=&\lim _{\epsilon \rightarrow 0}\frac{\log E_{{\mathcal P}}[m(\epsilon : {\mathcal P})]}{\log (1/\epsilon )}\\
&\geq &\lim _{\epsilon \rightarrow 0}\frac{E_{{\mathcal P}}[\log m(\epsilon : {\mathcal P})]}{\log (1/\epsilon )}\\
&=&\sum_{i=1}^{s}p({\mathcal P}_{i})\lim _{\epsilon \rightarrow 0}\frac{\log m(\epsilon :{\mathcal P}_{i})}{\log (1/\epsilon )}\\
&=&\sum_{i=1}^{s}p({\mathcal P}_{i}) {\rm Ddim}({\mathcal P}_{i}).
\end{eqnarray*}
This completes the proof of Theorem \ref{fusionddim}.
\hspace*{\fill}$\Box$\\
\end{comment}
where we have used Jensen's inequality to derive the first inequality. 
 We call the lower bound (\ref{fusiondef}) the {\em pseudo Ddim for model fusion} ${\mathcal F}^{\odot}$.
In the rest of this paper, we adopt it as Ddim value for model fusion. We write it as ${\rm Ddim}({\mathcal F}^{\odot})$.
Model fusion  is a reasonable setting  when we consider the transition period of model changes.
Then Ddim is no longer integer-valued.

\section{Model Change Sign Detection}

\begin{figure}[thb]
	%\vskip 0.2in
%\centering
%\begin{minipage}%{0.4\hsize}
	\begin{center}
%\centering
		%\centerline{
\includegraphics[width=70mm, height=45mm
%width=\hsize %height=50mm
]{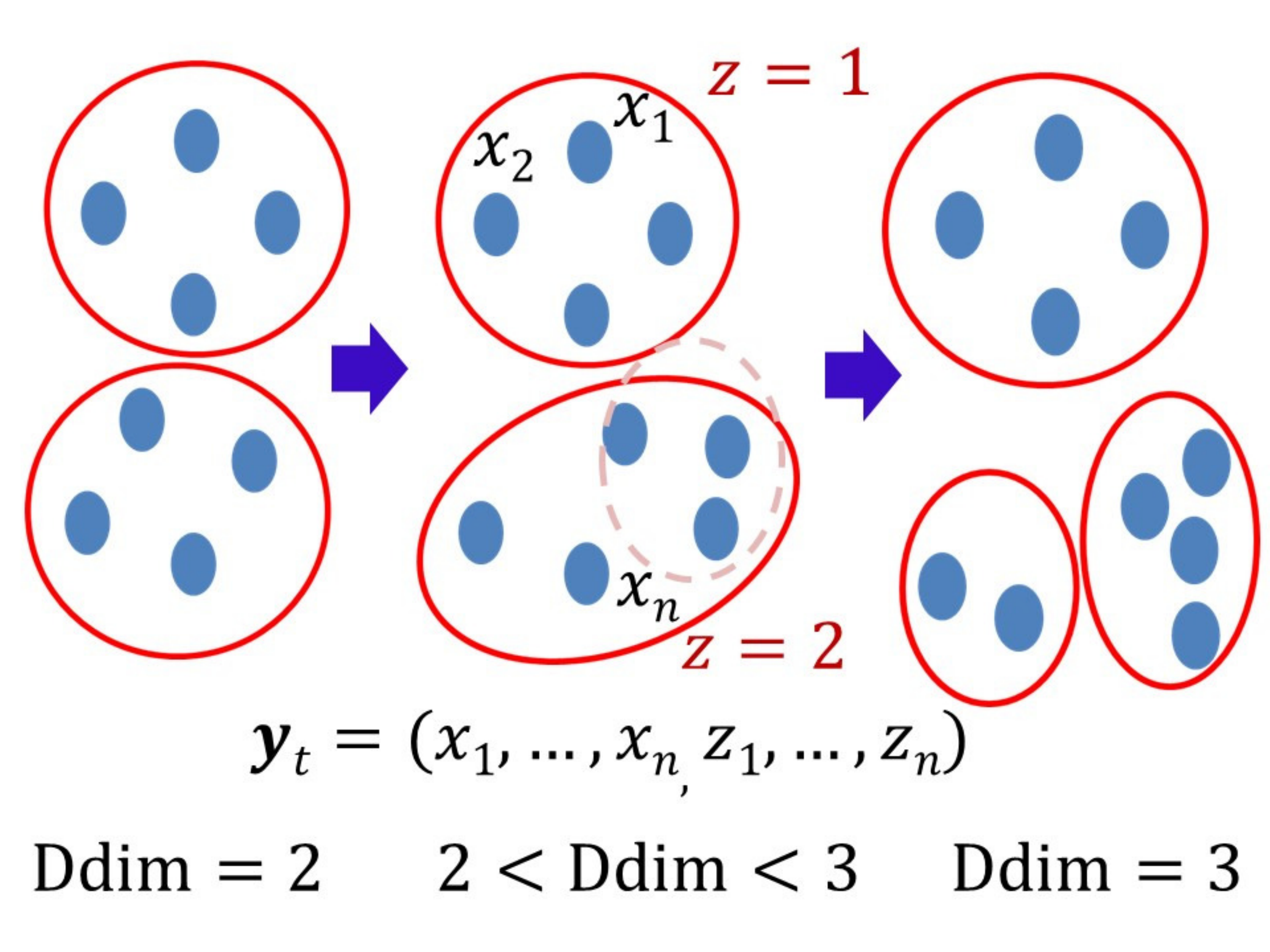} %}% windowsize_auc-eps-converted-to.pdf}}
		\caption{Continuous Model Selection}
		\label{changesign}
%\vspace*{-0.5cm}
	\end{center}
%\end{minipage}
\end{figure}

\subsection{Continuous Model Selection for GMMs}

This section proposes a methodology for detecting signs of model changes with continuous model selection.
We first focus on the case where the model is a {\em Gaussian mixture model}~(GMM).
The problem setting is as follows:
At each time we obtain a number of unlabeled multi-dimensional examples. By observing such examples sequentially, we obtain a data stream of the examples.
At each time, we may conduct clustering of the examples using GMMs.
Assuming that the number of components in GMM may change over time, we aim at detecting their changes and the signs of them.

The key idea is to conduct {\em continuous model selection}, which is to determine the real-valued model dimensionality on the basis of Ddim.
Below we give a scenario of continuous model selection with applications to model change sign detection.
Let ${\mathcal P}_{k}$ be a class of GMMs with $k$ components.
We consider the situation where the structure of GMM gradually changes over time (Fig. \ref{changesign}), while the model $k$ may abruptly change. 
The key observation is that
during the model transition period, model fusion occurs where a number of GMMs with different $k$s are probabilistically mixed according to the posterior probability distribution for  the given data. {Then model dimensionality in the transition period can be calculated as Ddim of model fusion. Thus we can detect signs of model changes by tracking the rise-up of Ddim. }

Below let us formalize the above scenario.
Let ${\mathcal X}$ be an $m$-dimensional real-valued domain and let $x\in {\mathcal X}$ be an observed datum.
Let $z\in \{1,\dots , k\}$ be a latent variable indicating which component $x$ comes from. 
Let $\mu _{i}\in {\mathbb R}^{m}$, $\Sigma _{i}\in {\mathbb R}^{m\times m}$ 
be the mean vector and variance-covariance matrix  for the $i$th component, respectively.
Let $\sum _{i}\pi _{i}=1,\ \pi _{i}\geq 0\ (i=1,\dots ,k)$.
Let $\theta =(\mu _{i}, \Sigma _{i}, \pi _{i})\mid _{i=1,\dots ,k}$.
Then a complete variable model of GMM with $k$-components is given by
    \begin{eqnarray*}
      p( x,z; \theta ,k )&=&p(x\mid z; \mu, \Sigma)p(z; \pi ),
\end{eqnarray*}
where 
\begin{align}
&p( x\mid z=i; \mu, \Sigma)
=\frac{1}{(2\pi )^{\frac{m}{2}}\cdot \mid \Sigma _i \mid^{\frac{1}{2}}} \exp \left \{ -\frac{1}{2}  (x-\mu _i)^{\top} \Sigma _i^{-1} (x-\mu _i) \right \}, \nonumber \\
&p(z=i;\pi )=\pi _{i}\ \ (i=1, \dots , k). \label{gmmpara}
\end{align}

Let ${\bm x}=x_{1},\dots , x_{n}$ be a sequence of observed variables of length $n$.
Let $z_{j}$ denote a latent variable which corresponds to $x_{j}$ and  ${\bm z}=z_{1},\dots , z_{n}$. 
Let ${\bm y}=({\bm x},{\bm z})$ be a complete variable.
Let $\hat{\mu}_{i}, \hat{\Sigma} _{i}$ be the maximum likelihood estimators of $\mu _{i}, \Sigma_{i}$ $(i=1,\dots ,k)$ for given 
${\bm y}$. 
Let $\hat{\pi}_{i}=n_{i}/n$ where $n_{i}$ is the number of occurrences in $\hat{z}^{n}$ 
such that $z=i$ $(i=1,\dots ,k)$ and $\sum ^{k}_{i=1}n_{i}=n$.
 ${\bm z}$ may be estimated by sampling from the posterior probability obtained by the EM algorithm.  
Let $\hat{\theta}({\bm y})=(\hat{\pi}_{i},\hat{\mu} _{i},\hat{\Sigma} _{i})\mid_{i=1,\dots ,k}$.

The NML codelength of ${\bm y}$ for a complete variable model of a GMM is given by 
\begin{eqnarray}\label{nmlupper}
L_{_{\rm NML}}({\bm y};k)&=&-\log p_{_{\rm NML}}({\bm y};k) \nonumber \\
      &=& -\log p({\bm y}; \hat{\theta}({\bm y}) ,k)
       + \log \mathcal{C}_{n}(k), %\label{eqn:uNMLcodeGauMix} 
\end{eqnarray}
where $\mathcal{C}_{n}(k)$ is 
a parametric complexity for a GMM.
According to \cite{correction}, an upper bound on $\mathcal{C}_{n}(k)$ is given as follows:
\begin{eqnarray}\label{pcomp}
\mathcal{C}_{n}(k) 
 %     &\buildrel \rm def \over =& \sum_{h_1,\cdots,h_K} \frac{N!}{h_1!\cdot \cdots \cdot h_K!} \prod_{k=1}^K \left( \frac{h_k}{N} \right)^{h_k} \mathcal{C}(h_k) \\
     &  \leq& \sum_{n_1,\cdots, n_k} \frac{n!}{n_1! \cdots  n_k !} \times  \prod_{i=1}^k \left( \frac{n_i}{n} \right)^{n_i}
      B(m,R,\epsilon) \nonumber \\
      & & \ \ \ \times \left( \frac{n_i}{2{\rm e}} \right)^{\frac{mn_i}{2}} \left( \Gamma_ m\left(\frac{n_i-1}{2}\right)\right)^{-1}, 
\end{eqnarray}
where
\begin{eqnarray}
\label{pcomp2}
& &B(m,R,\epsilon) \buildrel \rm def \over = \frac{2^{m+1}R^{\frac{m}{2}} %\prod_{d=1}^m {\epsilon_{d}}
\epsilon ^{-\frac{m^{2}}{2}}} {m^{m+1}\cdot \Gamma \left( \frac{m}{2} \right)}, \nonumber
    \end{eqnarray}
where $R$ is a positive constant such that for all $i$, 
$|| \hat{\mu}_{i}||^{2}\leq R$ and $\epsilon$ is a positive constant such that %$\epsilon _{d}$ 
$\epsilon$ is the lower bound on 
the %$d$-th 
smallest eigenvalue of $\Sigma _{i}$ for any $i$. 
$\Gamma _{m}$ is the multivariate Gamma function defined as 
$\Gamma _{m}(x)=\pi ^{\frac{m(m-1)}{4}}\prod ^{m}_{j=1}\Gamma (x+\frac{1-j}{2})$ and $\Gamma$ is the Gamma function.
We use the bound (\ref{pcomp}) as the value of ${\mathcal C}_{n}(k)$. 
It is known \cite{hirai} that $C_{n}(k)$ is computable in time $O(n^{2}k)$. 

At each time $t$, we observe a data sequence: ${\bm x}_{t} =x_{1},\dots , x_{n}\in {\mathcal X}^{n}$ of length $n$. 
We sequentially observe such a datum  as show in Fig. \ref{changesign}.  
Let ${\bm x}^{T}={\bm x}_{1},\dots ,{\bm x}_{T}\ ({\bm x}_{t}\in {\mathcal X}^{n},\ t=1,\dots ,T)$ be an observed data sequence.
The length $n$ may vary over time.
We denote the joint sequence of observed variables and latent variables at time $t$ as ${\bm y}_{t}=({\bm x}_{t},{\bm z}_{t})$.

We suppose that a number of GMMs with different $k$s are fused according to the probability distribution $p(k\mid {\bm y}_{t})$ at each time.
We define $p(k\mid {\bm y}_{t})$ as the annealed posterior probability of $k$ for  ${\bm y}_{t}$: 
\begin{align}\label{pos99}
&p(k\mid {\bm y}_{t})
 \buildrel \rm def \over =\frac{(p_{_{\rm NML}}({\bm y}_{t};k)p(k\mid k_{t-1}))^{\beta}}{\sum _{k'}(p_{_{\rm NML}}({\bm y}_{t};k')p(k\mid k_{t-1}))^{\beta}} \nonumber \\
& =
\frac{\exp (-\beta L_{_{\rm NML}}({\bm x}_{t},{\bm z}_{t};k)+\beta \log p(k\mid k_{t-1}))}{\sum _{k'}\exp (-\beta L_{_{\rm NML}}({\bm x}_{t},{\bm z}_{t};k')+\beta \log p(k'\mid k_{t-1}))},
\end{align}
where 
$k_{t-1}$ is the dimensionality estimated at 
time $t-1$, and 
\begin{align}
p (k \mid k_{t-1})     \buildrel \rm def \over = \left\{
        \begin{array}{ll}
          1-\gamma & \mbox{if}~ k=k_{t-1}~{\rm and}~k_{t-1} \neq 1,k_{\rm max},\\
          1-\gamma /2~& \mbox{if}~ k=k_{t-1}~{\rm and}~k_{t-1} = 1,k_{\rm max},\\
          \gamma /2~ &\mbox{if}~ k=k_{t-1} \pm 1. \label{alpha} 
        \end{array}\right.  %\nonumber \\
      \end{align}
$\gamma (0<\gamma <1)$ is a parameter. We estimate $\gamma$ using the MAP estimator with the beta distribution $\mathrm{Beta}(a,b)$ being the prior as follows:  \[\hat{\gamma} =\frac{N_t+a-1}{t+a+b-2},\] where $N_t$ shows how many times the number of clusters has changed until time $t-1$,and $\hat{\alpha}$ .
In the experiments to follow, we set $(a,b)=(2,10)$.
%$\lambda_{\mathrm{SDMS}} = 10$.
$\beta (>0)$ is the temperature parameter. 
In our experiments to follow, we set  $\beta$ as
\begin{eqnarray}\label{temp}
\beta =1/\sqrt{n}.
\end{eqnarray}
This is due to the PAC-Bayesian argument~\cite{gibbs} for Gibbs posteriors. 

Note that (\ref{pos99}) is calculated on the basis of the NML distribution.  
This is because 
the probability distribution with unknown parameters should be estimated as the NML distribution  since it is the optimal distribution in terms  of the minimax regret (see Sec.2.1). 

By (\ref{fusiondef}),  
we can calculate Ddim of model fusion of GMMs with various $k$s at time $t$ as 
\begin{eqnarray}\label{ggraph}
{\rm Ddim}({\mathcal F}^{\odot})_{t}=
\sum _{k}p_{t}({\mathcal P}_k){\rm Ddim}({\mathcal P}_{k}), 
%\nonumber %\sum _{k\in {\mathcal K}_{t}}kp(k|{\bm x}(t))$
\end{eqnarray}
where $p_{t}({\mathcal P}_{k})$ is the probability of ${\mathcal P}_{k}$ at time $t$ and $p_{t}({\mathcal P}_{k})=p(k\mid {\bm y}_{t})$ in this case. 
Note that Ddim for GMM with $k$ components is $k(m^{2}/2+(5m/2))-1\approx kf(m)$ where $f(m)=m^{2}/2+(5m/2)$.
However,  in order to focus on the mixture size, we divide the true Ddim by $f(m)$ to
consider an alternative Ddim of the form of (\ref{gggraph}).
\begin{eqnarray}\label{gggraph}
\overline{{\rm Ddim}}({\mathcal F}^{\odot})_{t}
&\buildrel \rm def \over =&\sum _{k}p(k\mid {\bm y}_{t}){\rm Ddim}({\mathcal P}_{k})/f(m) \nonumber \\
& \approx  &\sum _{k}p(k\mid {\bm y}_{t})k.
\end{eqnarray}
The calculation of real-valued $k$ according to (\ref{gggraph}) 
is really continuous model selection.

Suppose that there exists a true parametric dimensionality $k^{*}$. 
Then because of the {\em consistency} of MDL model estimation  (\cite{rissanen}, pp:63-69),
\[ p(\hat{k}=k^{*}\mid {\bm y}_{t}) \rightarrow 1\]
 for $\hat{k}$ minimizing the NML codelength as $n$ increases, 
Hence (\ref{gggraph}) will coincide with $k^{*}$ with probability $1$ as $n$ goes to infinity.
This implies that (\ref{gggraph}) is a natural extension of parametric dimensionality.

%The above methodology for computing Ddim is not restricted to GMM, but rather is applicable to general parametric classes as shown in Section 4.2.

\subsection{Model Change Sign Detection Algorithms}
Consider the situation where we sequentially observe a complete variable sequence: ${\bm y}_{1}, {\bm y}_{2},\dots ,{\bm y}_{T}$.
We then obtain a Ddim graph: 
\[\{(t, \overline{{\rm Ddim}}({\mathcal F}^{\odot})_{t}): t=1,2,\dots, T\},\]  
as in Fig.~\ref{ddim}. 
We can visualize the transition period by drawing the Ddim graph versus time. 
In this paper we have defined a sign of a model change as the starting point of the latent gradual change associated with it.
Thus we can detect signs of model changes by looking at the 
rise-up of Ddim. 

More precisely, we propose the following two methods for raising alarms of model change signs.

{\em 1) Thresholding method} (TH): We raise an alarm if 
the absolute difference between Ddim and 
the baseline 
exceeds a given threshold $\delta _{1}$.
The baseline is the parametric dimensionality estimated by 
the {\em sequential dynamic model selection algorithm} (SDMS)~\cite{hirai}, which is a sequential variant of DMS in \cite{ym05, ym07}. 
It outputs a model $k=\hat{k}$ with the shortest codelength, i.e., for $\lambda >0$, 
\begin{equation}\label{sdms}
\hat{k}=\argmin_{k}\{L_{_{\rm NML}}({\bm y}_{t};k)-\lambda \log p(k\mid k_{t-1})\},
\end{equation}
where $L_{_{\rm NML}}({\bm y}_{t};k)$ is calculated as in (\ref{nmlupper}). 
Letting $\hat{k}$ be the output of SDMS, we raise an alarm if 
\begin{eqnarray}\label{th}
{\rm TH}{\small -}{\rm Score}\buildrel \rm def \over =\mid \overline{{\rm Ddim} }-\hat{k}\mid > \delta _{1}.
\end{eqnarray}

{\em 2) Differential method} (Diff): We raise an alarm if 
the time difference of Ddim exceeds a given threshold $\delta _{2}$. That is, letting $\overline{{\rm Ddim}}_{t}$ be Ddim at time $t$, 
then we raise an alarm if
\begin{eqnarray}\label{diff}
{\rm Diff}{\small -}{\rm Score}\buildrel \rm def \over =\mid \overline{{\rm Ddim}}_{t}-\overline{{\rm Ddim}}_{t-1}\mid > \delta _{2}.
\end{eqnarray}

The computational complexity of TH and Diff at each time $t$ 
is governed by that for computing the NML codelength (\ref{nmlupper}).  The first term in (\ref{nmlupper}) is computable in time $O(nk)$.
The second term in (\ref{nmlupper}) is computable in time $O(n^{2}k)$~\cite{correction}, but it does not depend on data, hence can be calculated for various $n$ and $k$ beforehand. It can be referred when necessary. Hence the computational complexity of 
TH and Diff at each time is $O(nK)$ 
 where $K$ is an upper bound on $k$. 

\subsection{Continuous Model Selection for AR model}

The above methodology can also be applied to general classes of finite mixture models other than GMMs.
It can also be applied to general classes of parametric probabilistic model classes. We illustrate the case of {\em AR~(auto-regression)} model as an example. This model does not include latent variables.

Let a data sequence $\{x_{t}\},\ x_t\in {\mathbb R}\ (t=1,2,\dots ,T)$ be given.
For the modeling of the data sequence, we consider $AR(k)$~($k$-th order auto-regression) model of the form: 
\[x_{t}=a_{1}x_{t-1}+\cdots + a_{k}x_{t-k}+\epsilon ,\]
where $a_{i}\in {\mathbb R}\ (i=1,\dots ,k)$ are unknown parameters, and $\epsilon $ is a random variable following the Gaussian distribution with mean $0$ and unknown variance $\sigma ^{2}$. We set $\theta =(a_{1},\dots ,a_{k},\sigma ^{2})$.

Let ${\bm x}_{t}=x_{t},\dots , x_{t-w+1}$ be the $t$-th session for a window size $w$.  
We calculate the NML codelength $L_{_{\rm NML}}({\bm x}_{t}; k)$ for ${\bm x}_{t}$ associated with $AR(k)$ in the following sequential manner:
\begin{eqnarray*}
L_{_{\rm NML}}({\bm x}_{t}; k)=\sum ^{t}_{j=t-w+1}
\left( -\log \frac{p(x_{j}; \hat{\theta}(x_{j},x^{j-1}))}{\int p(y_{j}; \hat{\theta}(y,x^{j-1}))dy} \right).
\end{eqnarray*}

Letting ${\mathcal P}_{k}=AR(k)$, similarly with (\ref{pos99}) and (\ref{ggraph}), we can calculate Ddim at time $t$ and draw a Ddim graph.
%by
%\begin{eqnarray*}
%{\rm Ddim}({\mathcal F}^{\odot})_{t}=\sum _{k}p_{t}({\mathcal P}_{k})
%{\rm Ddim}({\mathcal P}_{k})=\sum _{k}p_{t}({\mathcal P}_{k})k,
%\end{eqnarray*}
%where 
%\begin{eqnarray*}
%p_{t}({\mathcal P}_{k})=\frac{\exp (-\beta L_{_{\rm NML}}({\bm x}_{t};k))p(k|k_{t-1})}{\sum _{k'}\exp (-\beta L_{_{\rm NML}}({\bm x}_{t};k'))p(k'|k_{t-1})}.
%\end{eqnarray*}
%Then we can draw the graph
%\[\{(t, {\rm Ddim}({\mathcal F}^{\odot})_{t}): t=1,2,\dots, T\}.\]  
We thereby possibly detect model change signs by applying TH and Diff to the graph.

\section{Experimental Results}
\subsection{Synthetic Data: GMM}

\subsubsection{Data set}
We employ synthetic data sets to evaluate how well we are able to detect signs of model changes using Ddim.
We let $n=1000$ at each time. 
We generated DataSet 1 according to GMMs so that the number of components changed from $k=2$ to $k=3$ as follows:
\begin{eqnarray}
       % p(\mathcal{P}): 
        \begin{cases}
        %Prob(K=2) = 1
        k=2,\  \mu =(\mu _{1},\mu_{2}) & {\rm if} \ 0\leq t \leq \tau_1, \nonumber \\
        %Prob(K=3) = 1
        k=3, \ \mu =(\mu_1, \mu_2, f_{\alpha}(t))
        %\frac{(\tau_2-t)^{\alpha}\mu_2 + (t-\tau_1)^{\alpha}\mu_3}{(\tau_2-t )^{\alpha}-(t-\tau_1)^{\alpha}} )
        & {\rm if} \tau_1+1 \leq t \leq \tau_2, \nonumber \\
        %Prob(K=3) = 1
        k=3, \ \mu =(\mu_1, \mu_2, \mu_3)& {\rm if} \ \tau_2 +1 \leq t \leq T, 
        \end{cases} \label{changesim} 
        \end{eqnarray}
where 
\begin{eqnarray}
f_{\alpha}(t)\buildrel \rm def \over = \frac{(\tau_2-t)^{\alpha}\mu_2 + (t-\tau_1)^{\alpha}\mu_3}{(\tau_2-t )^{\alpha}+(t-\tau_1)^{\alpha}} \ \ (0< \alpha \leq 1). \label{al}
\end{eqnarray}
In it, one component collapsed gradually in the transition period from $t=\tau_1+1$ to $t=\tau_2$. 
$f_{\alpha}(t)$ is the mean value which switches from $\mu _{2}$ to $\mu _{3}$ where the speed of change is specified by a parameter $\alpha$. 
Fig. \ref{alphaq} shows %the normalized graph of $f_{\alpha}(t)$
the graph of the proportion of $\mu_{3}$ in the mean to $\mu_{2}$ 
for various $\alpha$.
The change becomes rapid as $\alpha $ approaches to zero. 
The variance covariance matrix of each component is given by 
\begin{eqnarray}\label{variance}
\Sigma =(rAA^{{\rm T}}+(I-r)I)\times {\rm var},
\end{eqnarray}
where $r=0.2$, ${\rm var}=3$ and $A$ is a randomly generated $m\times m$ matrix.
We set $ m=3, \tau_1=9,\tau_2=29,T=39$.
It appears that the number of components of a GMM abruptly changed at $t=20$ since it takes a discrete value. However, in the early stage of $k=3$, the model is very close to $k=2$ because the mean values of Gaussian components are very close each other. 
It may be more natural to recognize the model dimensionality at this stage as a value between $k=2$ and $k=3$.

\begin{figure}[!htb]
%\vspace*{-2cm}
\begin{center}
%\hspace*{-4.0cm}
          %%% result for data set 1 for structural entropy
          \includegraphics[keepaspectratio, width=10.0cm]{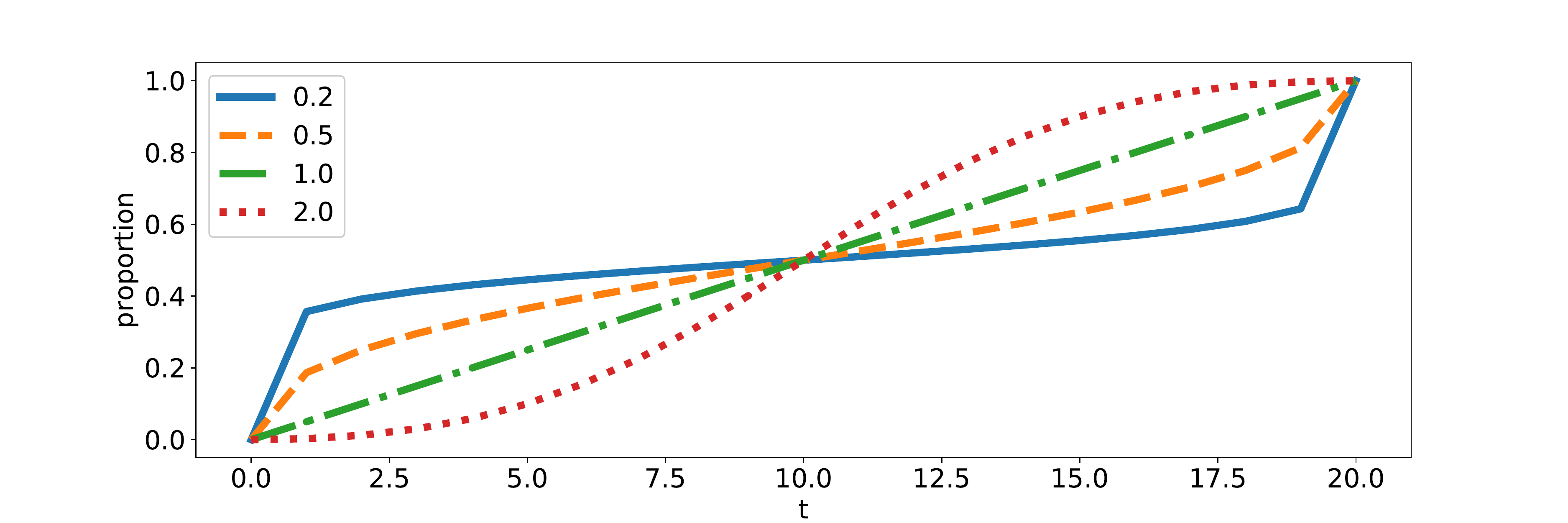}
          %./jpg/ddim/single_variation_alpha/
            %height=3.0cm]
%\ \ \\
%\ \ \\
%\ \ \\
%\ \ \\
%\ \ \\
%\ \ \\
         \caption{Mean change in transition period for various $\alpha$}\label{alphaq}

\end{center}
\end{figure}

We evaluate how well Ddim tracked the transition period of model change. %The prior distribution for GMMs was the uniform distribution. 
The temperature parameter $\beta$ was chosen so that $\beta =0.0316$ according to (\ref{temp}).
Fig. \ref{f1}
%\ref{fig:gmm_clu_ddim_01} 
shows how Ddim gradually grows as time goes by for various $\alpha$ values. 
The gray zone shows the transition period when the model changes from $k=2$ to $k=3$.
The blue line shows the number of components of the GMM  estimated by the SDMS algorithm as in (\ref{sdms}). 
The green curve shows the Ddim graph.
The red and purple curves show TH-Score and Diff-Score as in (\ref{th}) and (\ref{diff}), respectively.  We show the time points of their alarms using the same colors.
Ddim successfully visualized how rapidly the GMM structure changed in the transition period from $t=10$ to $29$. 
The true change occurs rapidly for $\alpha=0.2$, while it occurs slowly for $\alpha =1.0$. Ddim was able 
to successfully track their transition process depending on $\alpha$. Ddim detected signs earlier than SDMS made an alarm of model change.
\begin{figure}[htb]
\begin{center}
\begin{tabular}{c}
\begin{minipage}{0.5\hsize}
%\hspace*{-3.0cm}
\includegraphics[keepaspectratio, width=7cm]
%{./experiment20190123/ddim_single/gmm_clu_se_ddim_single_rand0_alpha0.20-eps-converted-to.pdf}
{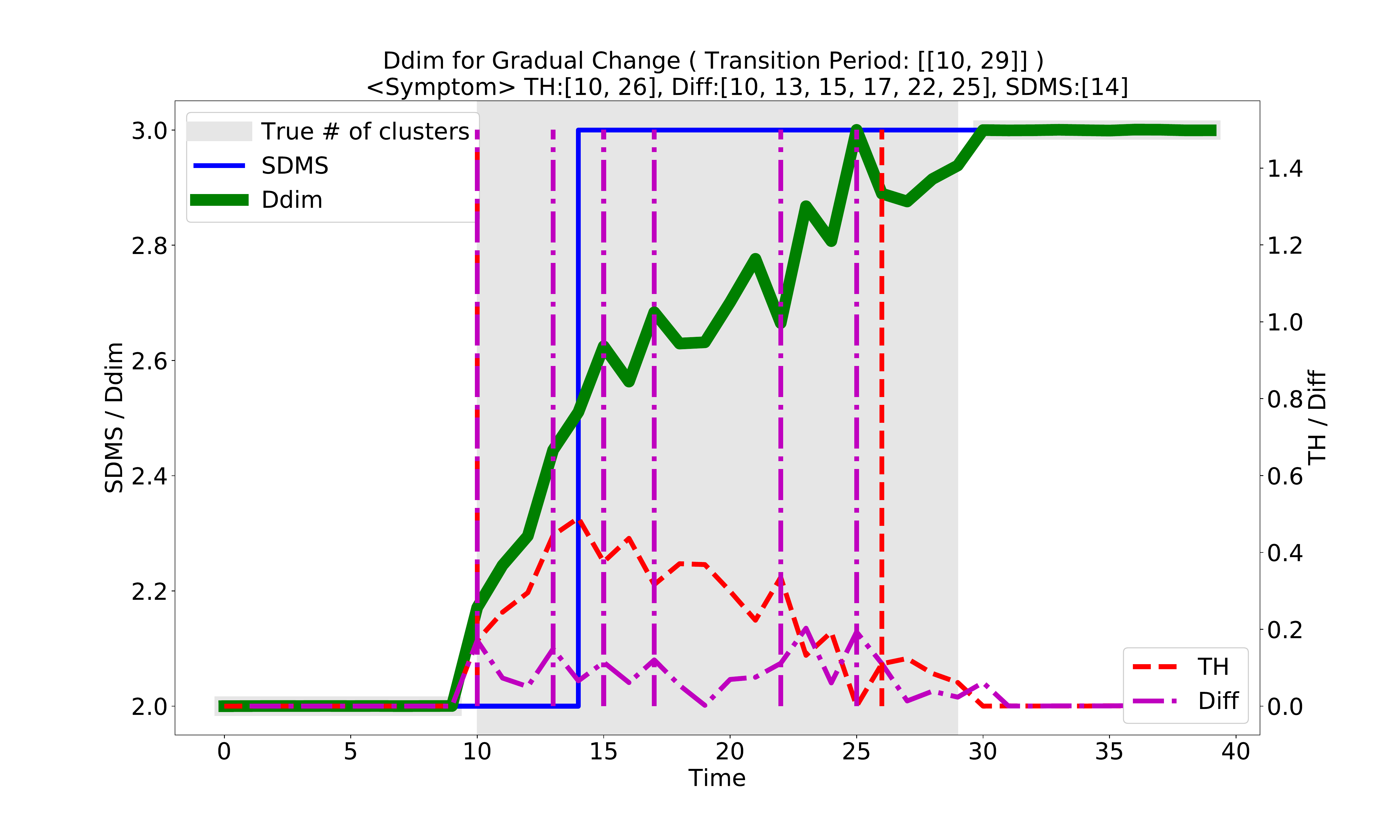}
% \ \ \\
%                    \vspace*{-0.5cm}
         \begin{center} \ \ \ $\alpha =0.2$
          \end{center}
          %{./jpg/ddim/gmm_clu_se_ddim_beta0.03-eps-converted-to.pdf}\ \ \\
\end{minipage}\\ \\
\begin{minipage}{0.5\hsize}
     %  \vspace*{-1.5cm}
      % \hspace*{-3.0cm}
        %\begin{center}
          %%% result for data set 1 for structural entropy
          \includegraphics[keepaspectratio, width=7cm] %height=3.0cm]
          {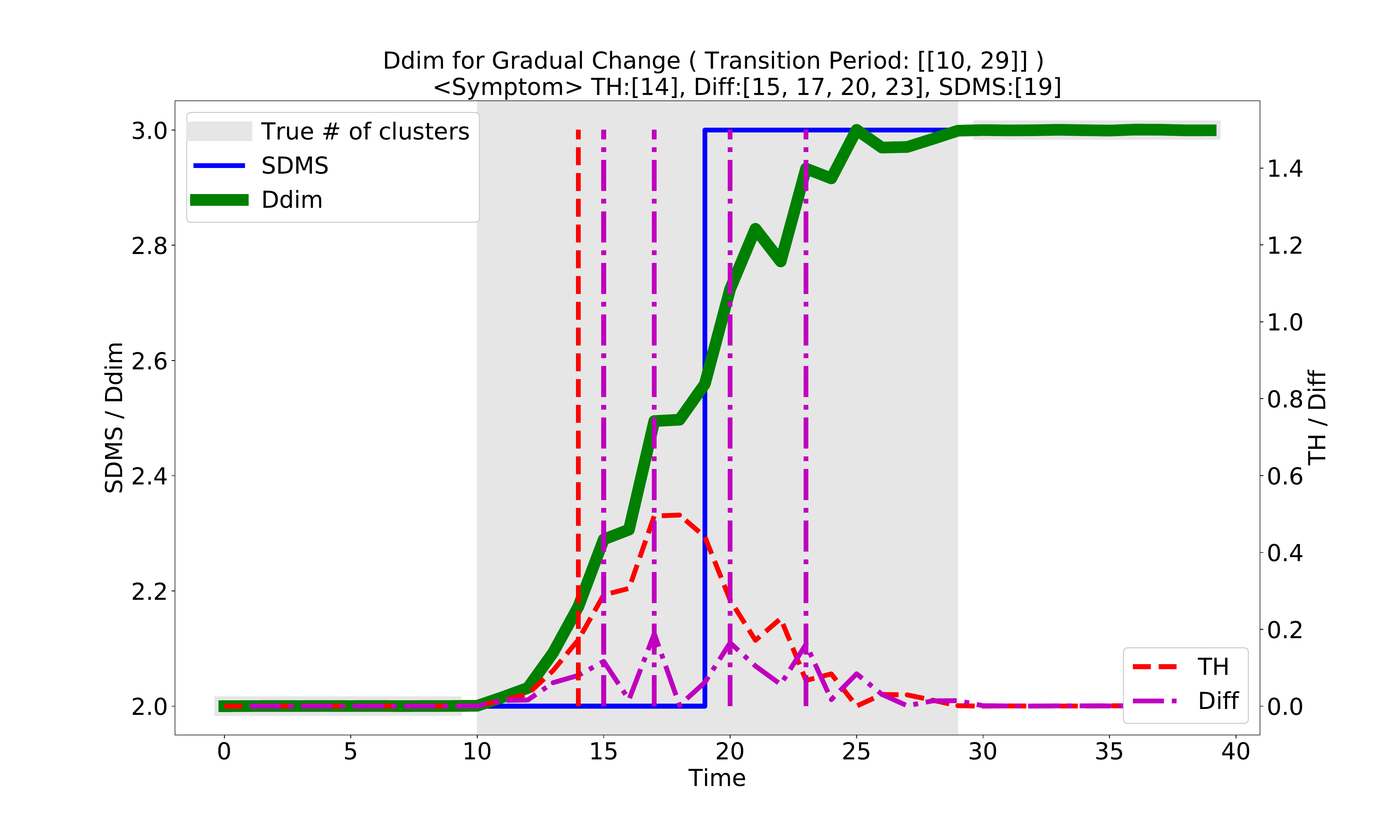}
%\ \ \\
          %{./jpg/ddim/gmm_clu_se_ddim_beta0.03-eps-converted-to.pdf}\ \ \\
       %             \vspace*{-0.5cm}
         \begin{center} $\alpha =0.5$
          \end{center}
\end{minipage}
\ \ \\
\ \ \\
\begin{minipage}{0.5\hsize}
\begin{center}
      % \vspace*{-1.5cm}
       %\hspace*{-3.0cm}
        %\begin{center}
          %%% result for data set 1 for structural entropy
          \includegraphics[keepaspectratio, width=7cm] %height=3.0cm]
          {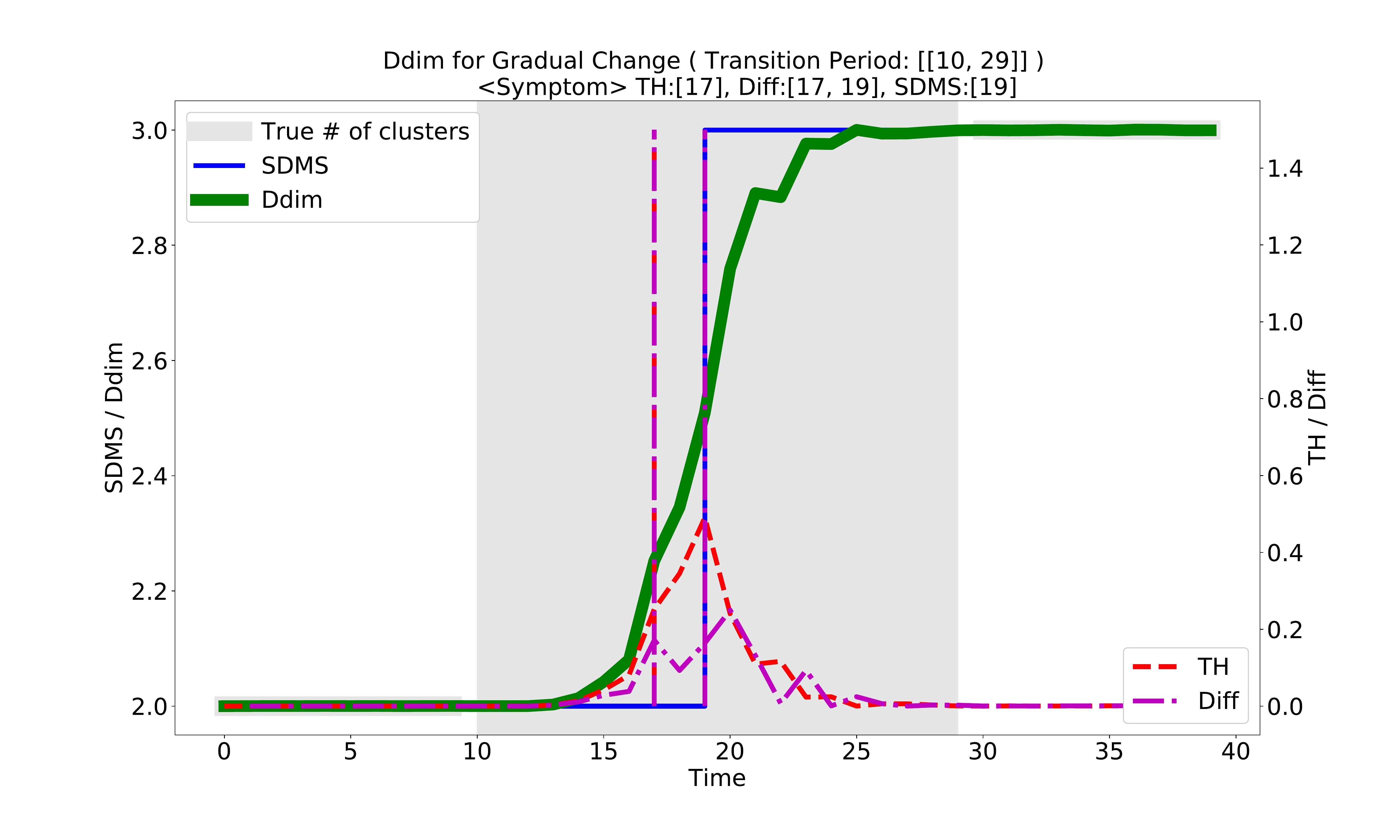}
          %./jpg/ddim/single_variation_alpha/%gmm_clu_se_ddim_rand0_alpha2.00-eps-converted-to.pdf
%\ \ \\
          %{./jpg/ddim/gmm_clu_se_ddim_beta0.03-eps-converted-to.pdf}\ \ \\
         %\vspace*{-0.8cm}
         \begin{center} $\alpha =1.0$
          \end{center}
          \end{center}
\end{minipage}
%          \ \ \\
         \ \ \\
\end{tabular}
%\vspace*{-0.3cm}
          \caption{Ddim graph %\& the estimated number of components
{\small  (transition period: $[\tau_1=9,\tau_2=29], T=39$)}}\label{f1}
        %\end{center}
%        \vspace*{-0.3cm}
\end{center}  
\end{figure}

\begin{figure}[!th]
       
        \begin{center}
       % \hspace*{-4.0cm}
        \centering
          %%% result for data set 1 for structural entropy
          \includegraphics[keepaspectratio, 
          width=8.0cm] %, height=4.2cm] %height=3.0cm]
          %{./jpg/ddim/multi/
{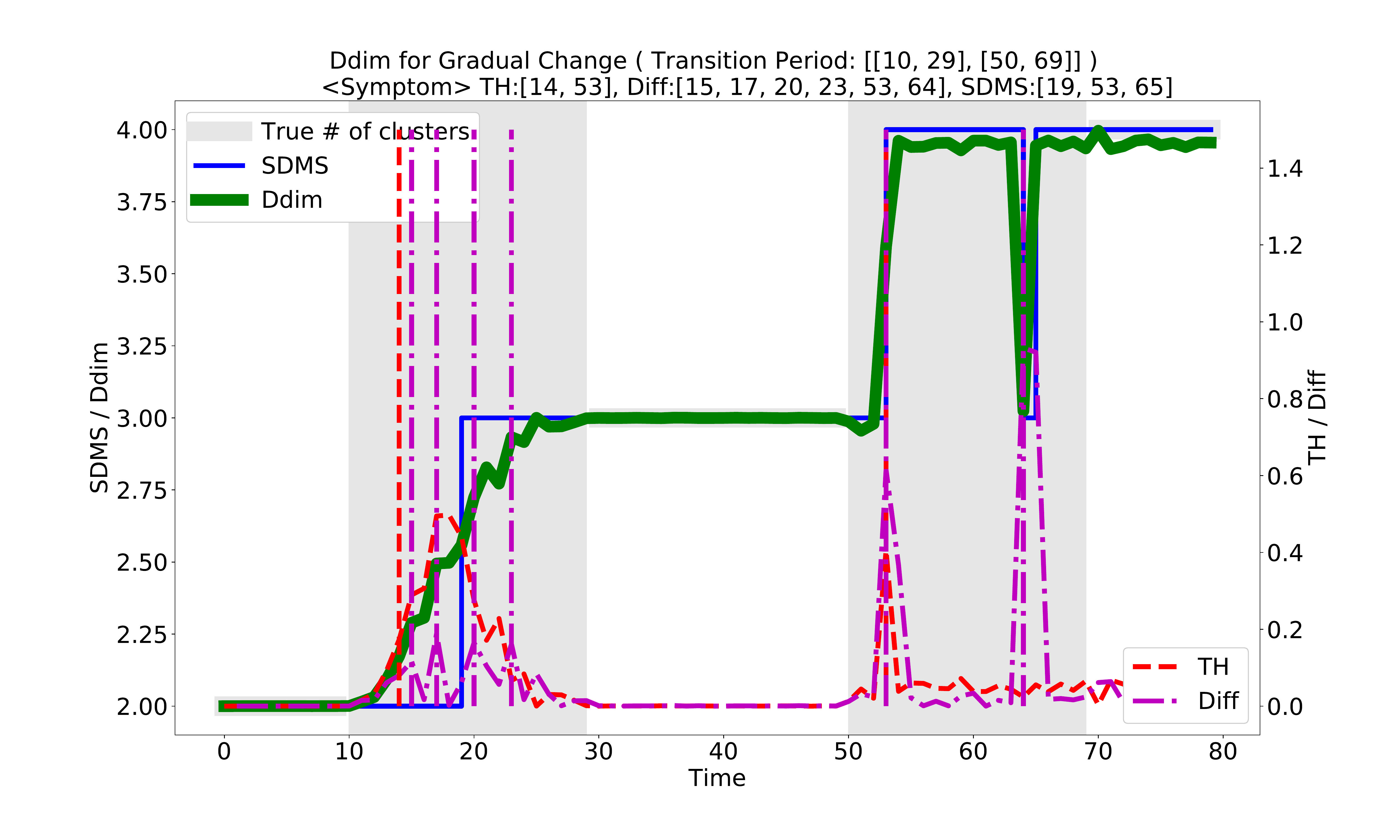}
%          \ \ \\
%          \ \ \\
%\vspace*{-0.7cm}
          \caption{Ddim graph %\& the estimated number of components 
{\small (transition periods: $[\tau_1=9,\tau_2=29],[\tau _3=49,\tau _{4}=59], T=79$)}}\label{multi}
        \end{center}
        \vspace*{-0.5cm}
      \end{figure}
      \ \ \\
      
We next consider the case where there are multiple change points. %of the number $k$ of components for GMM. 
We generated DataSet 2 according to GMMs so that the number of components changed from $k=2$ to $k=3$, then from $k=3$ to $k=4$.
as follows:
{\small
      \begin{eqnarray*}
       % p(\mathcal{P}): 
        \begin{cases}
        %Prob(K=2) = 1
        k=2,\  \mu =(\mu _{1},\mu_{2}) & {\rm if} \ 1\leq t \leq \tau_1, \\
        %Prob(K=3) = 1
        k=3, \ \mu =(\mu_1, \mu_2, \frac{(\tau_2-t)\mu_2 + (t-\tau_1)\mu_3}{\tau_2-\tau_1} )& {\rm if} \tau_1+1 \leq t \leq \tau_2,\\
        %Prob(K=3) = 1
        k=3, \ \mu =(\mu_1, \mu_2, \mu_3)& {\rm if} \ \tau_2 \leq t \leq \tau _{3}, \\
        k=4, \ \mu =(\mu_1, \mu_2, \mu _{3}, \frac{(\tau_4-t)\mu_3 + (t-\tau_4)\mu_4}{\tau_4-\tau_3} )& {\rm if} \tau_3+1 \leq t \leq \tau_4,\\
        k=4, \ \mu =(\mu_1, \mu_2, \mu_3, \mu_4)& {\rm if} \tau_4+1 \leq t \leq T. 
        \end{cases}
      \end{eqnarray*}
      }
One component collapsed gradually over time from $t=10$ to $t=29$ and the other one collapsed from $t=50$ to $t=69$. We set $T=79$.
In the transition periods the parameters varied as with the single change point case.

Fig. \ref{multi} shows the Ddim graph for $\alpha =0.5$.
The gray zone shows the transition periods when the model changes from $k=2$ to $k=3$ and $k=3$ to $k=4$.
The green curve shows the Ddim graph.
%The other colored lines have the same meanings as in Fig.~\ref{f1}.
The blue line shows the number of components of the GMM estimated by SDMS.
The red and purple lines show the times when alarms for signs of model changes are raised using TH and Diff, respectively. 
The Ddim graph helps us understand well how rapidly the GMM structure gradually changes in the transition periods from $t=10$ to $t=29$ and from $t=50$ to $t=69$.
%The Ddim graph helps us  understand well how rapidly the model changes in the transition period.
TH detected the signs of model changes earlier than SDMS.

\begin{table}[!t]
\begin{comment}
\begin{center}
\caption{Benefit comparison results} \label{tb1}
(a) Single change point\\
%\begin{tabular}{|ccccc|}
% \multicolumn{2}{c}{} & 
\begin{tabular}{|rrrrr|} \hline
Methods & $\alpha=$0.2 & $\alpha=$0.5 & $\alpha=$1.0 & $\alpha=$2.0 \\ \hline 
 TH & {\bf 0.97 $\pm$ 0.04} & {\bf 0.79 $\pm$ 0.03} & {\bf 0.67 $\pm$ 0.02} & {\bf 0.59 $\pm$ 0.02} \\ 
 Diff & 0.96 $\pm$ 0.07 & 0.74 $\pm$ 0.07 & 0.64 $\pm$ 0.05 & 0.58 $\pm$ 0.02 \\ 
 SDMS & 0.69 $\pm$ 0.16 & 0.59 $\pm$ 0.05 & 0.55 $\pm$ 0.04 & 0.51 $\pm$ 0.02 \\ 
 FS & 0.52 $\pm$ 0.16 & 0.42 $\pm$ 0.06 & 0.42 $\pm$ 0.05 & 0.42 $\pm$ 0.03 \\ 
 FSW-TH & 0.79 $\pm$ 0.08 & 0.62 $\pm$ 0.05 & 0.55 $\pm$ 0.04 & 0.51 $\pm$ 0.02 \\ 
 FSW-Diff & 0.61 $\pm$ 0.16 & 0.48 $\pm$ 0.07 & 0.47 $\pm$ 0.04 & 0.52 $\pm$ 0.11 \\ 
 \hline \end{tabular}

\ \ \\
(b) Multiple change points\\
%\begin{tabular}{|ccccc|}
% \multicolumn{2}{c}{} & 
\begin{tabular}{|rrrrr|} \hline
Methods & $\alpha=$0.2 & $\alpha=$0.5 & $\alpha=$1.0 & $\alpha=$2.0 \\ \hline 
TH & {\bf 0.98 $\pm$ 0.02} & {\bf 0.82 $\pm$ 0.04} & {\bf 0.73 $\pm$ 0.06} & 0.61 $\pm$ 0.04 \\ 
Diff & 0.98 $\pm$ 0.03 & 0.81 $\pm$ 0.03 & 0.72 $\pm$ 0.06 & 0.66 $\pm$ 0.08 \\ 
SDMS & 0.81 $\pm$ 0.13 & 0.66 $\pm$ 0.08 & 0.61 $\pm$ 0.04 & 0.55 $\pm$ 0.03 \\ 
FS & 0.72 $\pm$ 0.09 & 0.59 $\pm$ 0.06 & 0.53 $\pm$ 0.03 & 0.47 $\pm$ 0.02 \\ 
FSW-TH & 0.89 $\pm$ 0.04 & 0.75 $\pm$ 0.03 & 0.63 $\pm$ 0.02 & 0.56 $\pm$ 0.02 \\ 
FSW-Diff & 0.78 $\pm$ 0.10 & 0.67 $\pm$ 0.08 & 0.65 $\pm$ 0.09 & {\bf 0.67 $\pm$ 0.10} \\ \hline \end{tabular}

\end{center}
\end{comment}

\begin{center}
\caption{AUC comparison results for GMMs} \label{tb1}
%\begin{minipage}{0.40\hsize}
\begin{center}
(a) Single change point\\
%\begin{tabular}{|ccccc|}
% \multicolumn{2}{c}{} & 
\begin{tabular}{|rrrrr|} \hline
Method & $\alpha=$0.2 & $\alpha=$0.5 & $\alpha=$1.0 & $\alpha=$2.0 \\ \hline 
 TH &{\bf 0.995} & {0.920}& {0.845}& {\bf 0.802}\\ 
 Diff &{\bf 0.995} & 0.907&0.830 &0.797 \\ 
 SDMS &0.850 &0.797 & 0.775&0.757\\ 
 FS & 0.755 & 0.715& 0.710& 0.710 \\ 
 FSW-TH & 0.893& 0.813& 0.778 & 0.758\\ 
 FSW-Diff & 0.825 & 0.778& 0.750& 0.758 \\ 
{SE} & {{\bf 0.995}}& {{\bf 0.927}}& {{\bf 0.853}}& {0.801}\\
 \hline \end{tabular}
\end{center}
%\end{minipage}
\ \ \\
%\begin{minipage}{0.40\hsize}
\begin{center}
(b) Multiple change points\\
%\begin{tabular}{|ccccc|}
% \multicolumn{2}{c}{} & 
\begin{tabular}{|rrrrr|} \hline
Method & $\alpha=$0.2 & $\alpha=$0.5 & $\alpha=$1.0 & $\alpha=$2.0 \\ \hline 
TH &{\bf 0.998} & { 0.925}& {0.893} & {\bf 0.870} \\ 
Diff & {\bf 0.998}& 0.920& 0.883 &0.850 \\ 
SDMS &0.905 &0.831 & 0.804 &0.773 \\ 
FS &0.856 & 0.794& 0.761& 0.739\\ 
FSW-TH & 0.945&0.874 &0.816 & 0.780\\ 
FSW-Diff &0.899 &0.855 & 0.834 &0.835 \\ 
{SE} &{0.997} &{{\bf 0.928}} &{{\bf 0.903}} &{0.867}\\
\hline 
\end{tabular}
%\vspace*{-0.5cm}
\end{center}
%\end{minipage}
\end{center}
\end{table}

\subsubsection{Evaluation metrics}
Next we quantitatively evaluate how early we were able to detect signs of model changes with Ddim.
We measure the performance of any algorithm in terms of 
benefit. %and false alarm rate.
Let $\hat{t}$ be the first time when an alarm is made and $t^{*}$ be the true sign, which we define as {\em the starting point of model change}.
Then {\em benefit} is defined as 
\begin{eqnarray}\label{defbenefit}
{\rm benefit}=\begin{cases}
1-(\hat{t}-t^{*})/{U} & (t^{*}\leq \hat{t}<t^{*}+U),\\
0 & {\rm otherwise},
\end{cases}
\end{eqnarray}
where $U$ is a given parameter. Benefit takes the maximum value $1$ when the alarm coincides with the true sign. It decreases linearly as $t$ goes by and becomes zero as $\hat{t}$ exceeds $t^{*}+U$. 
%A false alarm may be defined as that raised outside the transition period. However, in all of the methods for comparison, there was no such an alarm. Thus we employed only benefit as a performance metric.
{\em False alarm rate}~(FAR) is defined as the ratio of the number of alarms outside the transition period over the total number of alarms.

%We evaluate TH~(\ref{th}) and Diff~(\ref{diff}) with threshold $\delta _{1}=0.1$ and $\delta _{2}=0.1$, respectively. It means that an alarm is raised when Ddim is $10\%$ different from the baseline.
We evaluate any method for model change sign detection algorithm in terms of Area Under Curve~(AUC) of Benefit-FAR curve that is obtained by varying the threshold parameter $\delta $ such as in (\ref{th}) and (\ref{diff}). 
We set $U=10$ in (\ref{defbenefit}).

\subsubsection{Methods for comparison}
\begin{comment}
\textcolor{red}{When we select methods for comparison,  we require the following two conditions:
A) They must detect signs of model changes, and B) they must measure the representation power of a model in the transition period.
We consider the following three methods for comparison satisfying the conditions A) and B). } \\
\end{comment}

We consider the following methods for comparison.\\
1) {\em The sequential DMS algorithm} (SDMS)~\cite{hirai}:
The SDMS algorithm with $\lambda =1$ outputs the estimated parametric dimensionality as in (\ref{sdms}). 
We raise an alarm when the output of SDMS changes.\\
2) {\em  Fixed share algorithm} (FS)~\cite{herbster}:
We think of each model $k$ as an {\em expert}, and perform Herbster and Warmuth's {\em fixed share algorithm}, abbreviated as FS. It was originally designed to make prediction by taking a weighted average over a number of experts, where the weight is calculated as a linear combination of the exponential update weight and the sum of other experts' ones. 
In it the expert with the largest weight is the best expert, which may change over time. 
We can think of FS as a model change detection 
algorithm by tracking the time-varying best expert.

Here is a summary of FS.
 Let $k$ be the index of the expert. $L({\bm z}_{t-1})$ is the loss function for the $k$th expert for data ${\bm z}_{t-1}$, which is the NML codelength in our setting.
$w_{t,k}^{u}$ and $w_{t,k}^{s}$ are tentative and final weights for the $k$th expert at time $t$. FS conducts the following weight update rule:  Letting $\alpha >0$ be a sharing parameter and $\beta$ be a learning ratio, 
   \begin{eqnarray*}
   \label{fsbeta}
      w_{t-1,k}^u &=& w_{t-1,k}^s \cdot \exp\{-\beta L_{k}({\bm z}_{t-1})\} , \\
      \label{fsalpha}
      w_{t,k}^s &=& (1-\alpha)w_{t-1,k}^u + \sum_{\ell\neq k} \frac{\alpha}{n-1}w_{t-1,\ell}^u .
      %\\
%      k_{_{\mathrm{FS}}}(t) &=& \sum_{k\in \{\hat{K}_t,\hat{K}'_t\}} k \cdot w_{t,k}^s.
    \end{eqnarray*}
 Let  $\hat{k}$  be the best expert in which $w_{t,k}^{s}$ is maximum.
 FS raises an alarm when the best expert changes. 
The learning rate was set to be the same as our method. \\
3) {\em Fixed share weighted algorithm} (FSW-TH, FSW-Diff): 
We consider variants of TH and Diff where $p(k\mid{\bm y}_{t})$ as in (\ref{pos99}) is replaced with the normalized weight  for $k$ calculated in the process of FS. FSW-TH and FSW-Diff calculate scores by plugging  $w_{t,k}^{s}$ to $p(k\mid {\bm y}_{t})$ in (\ref{gggraph}) and make alerts according to  (\ref{th}) and (\ref{diff}), respectively.
%It has no information-theoretic meanings for the weights. 
The learning rate was set to be the same as our method.
\\
{4) {\em Structural entropy} (SE):  It is a measure of uncertainty for model selection, developed in \cite{bigdata2018}. 
It is calculated as the entropy with respect to model according to the probability distribution (\ref{pos99}).
SE make alerts when it exceeds a threshold.}

The method 1) is the only existing work that performs on-line dynamic model selection. The methods 2) and 3) are the ones that can be adapted to our problem setting.
The method 1) and 2) are model change detection algorithms while the method 3) is an algorithm for quantifying latent gradual changes in a similar way with TH or DIFF. The method 4) is to detect  change signs from the view of model uncertainty, but not to intend continuous model selection.

\subsubsection{Results}
We generated random data $10$ times and took an average value of benefit over $10$ trials for each method.
Table \ref{tb1} shows results on comparison of all the methods in terms of AUC 
%benefit with standard deviation for various values of $\alpha$ as in (\ref{al}) 
both for single and multiple change cases.  
%AUC was calculated for the benefit-FAR curve obtained by varying a threshold. 
The parameter $\alpha$ specifies the speed of change.  
As for the multiple change cases, AUC was calculated as an average taken over all change points. 
Both for the single and multiple change cases,
TH and Diff had much higher benefit than the FS-based methods for all the cases. % except $\alpha =2$. 
It was statistically significant via t-test with p-values less than $5\%$.
This implies that TH and Diff were able to detect signs of model changes significantly earlier than the FS-based methods. TH worked almost as well as Diff. 
%Although FSW-Diff was better than TH for $\alpha =2$ in multiple changes, their difference was not statistically significant with p-values larger than $5\%$.
%The difference between TH(Diff) and FSW-Diff was not statistically significant with p-values larger than $5\%$ for $\alpha =2$ in multiple changes.

It is worthwhile noting that TH and Diff performed better than 
FSW-TH and FSW-Diff. It implies that the posterior based on the NML distribution is more suitable for tracking gradual model changes than that based on the FS-based heuristics. 
As $\alpha $ becomes small, the superiority of TH and Diff over the others becomes more remarkable. This implies that Ddim is able to catch up the growth of a cluster much more quickly  than the others. 

Both for the single and multiple change cases, TH worked slightly better than Diff, but they were almost comparable.
%For the single change case, TH worked slightly better than Diff when $\alpha $ was relatively small. 
%For the multiple change case, 
%TH worked better than Diff when $\alpha$ was close to zero, while they were comparable each other when $\alpha$ was getting large. 

%\textcolor{red}
{TH and Diff are comparable to SE in terms of change sign detection. Note that the Ddim based ones realize both model change sign detection and continuous model selection simultaneously, while SE can do only sign detection by measuring the uncertainty in model selection. 
Therefore, Ddim is effective in the sense that it is the only method that measures the representation power of a model as well as detects change signs.}

\subsection{Synthetic Data: Auto-regression model}
\subsubsection{Data sets}
We next examined continuous model selection for auto-regression (AR) models.
We let $n=1000$ at each time.
We generated DataSet 3 according to AR model in the setting as in Sec.3.3 where the number $k$ of coefficients in AR model changed over time as follows:
\begin{eqnarray*}
K=\begin{cases}
1  &{\rm if}\ 1\leq t\leq \tau _{1},\\
1 \  {\rm with\ prob.}\ 1-\frac{t-\tau _{1}}{\tau _{2}-\tau _{1}}& {\rm if}\ \tau_{1}\leq t\leq \tau _{2},\\
3\  {\rm with\ prob.}\ \frac{t-\tau _{1}}{\tau _{2}-\tau _{1}}& {\rm if}\ \tau_{1}\leq t\leq \tau _{2},\\
3 &{\rm if}\ \tau _{2}\leq t\leq T.
\end{cases}
\end{eqnarray*}

\subsubsection{Results}
We generated random data $10$ times and took an average value of benefit over $10$ trials for each method.
Table \ref{tb3} shows results on comparison of all the methods in terms of AUC for Benefit-FAR curves.
%benefit with standard deviation for various values of $\alpha$ as in (\ref{al}) both for single change and multiple change cases. 

\begin{table}[!h]
\begin{center}
\caption{AUC comparison results for AR models} \label{tb3}
\begin{tabular}{|rr|} \hline
Method & AUC \\ \hline 
 TH & {\bf 0.897} \\ 
 Diff & { 0.896} \\ 
 SDMS & 0.771 \\ 
 FS & 0.500 \\ 
 FSW-TH & 0.771  \\ 
 FSW-Diff & 0.500  \\
{SE} & {0.896 } \\
 \hline \end{tabular}
%\vspace*{-0.5cm}
\end{center}
\end{table}

Table \ref{tb3} shows that TH and Diff are comparable to SE and obtained much larger values of AUC than other methods except SE. 
This implies that TH and Diff could catch up signs of model changes significantly earlier than the others except SE.
{Note again that TH and Diff conduct not only change sign detection but also continuous model selection to understand the power of representation of a model in the transition period. Meanwhile SE only perform change sign detection.}

\subsection{Real Data: Market Data}

\subsubsection{Data sets}
We apply our method to real market data 
provided by 
HAKUHODO,INC. (https://www.hakuhodo-global.com/) and M-CUBE,INC. (https://www. m-cube.com/). 
This data set consists of 912 customers' beer purchase transactions from 
Nov. 1st 2010 to Jan. 31st 2011.
Each customer's
record is specified by a four-dimensional feature
vector, each component of which shows a consumption volume
for a certain beer category. Categories are: 
\{{Beer}(A), {Low-malt beer}(B), {Other brewed-alcohol}(C), {Liquor} (D)\}.

We constructed a sequence of customers' feature vectors as follows: 
A time unit is a day. At each time
$t(=\tau , . . . ,T)$, we denote the feature vector of the $i$th customer as $x_{it} = (x_{it,A}, . . . , x_{it,D}) \in {\mathbb R}^{4}$. Each $x_{it,j}$ is
the $i$th customer's consumption of the $j$th category from
time $t-\tau +1$ to $t$. We denote data at time $t$ as ${\bm x}_{t} = (x_{1t}, . . . , x_{nt})$, where $n=912$, the number of customers. 
The total number of transactions is $13993$. 
%$36480(=912\times 40)$.
%, the number of the customers.  
We set $\tau =14$ and $T=53$.
%who purchased the products from time t − τ + 1 to t
Since TH and DIFF have turned out to outperform the other methods in the previous section and we like to conduct continuous model selection simultaneously in order to visualize the dimension in the transition period, we focus on evaluating how well they work for the real data sets.

      \begin{figure}[!h]
        \begin{center}
      % \hspace*{-5cm}
          %%% result for data set 1 for structural entropy
          \includegraphics[keepaspectratio, 
          width=8cm] %, height=4.3cm] %height=3.0cm]
          {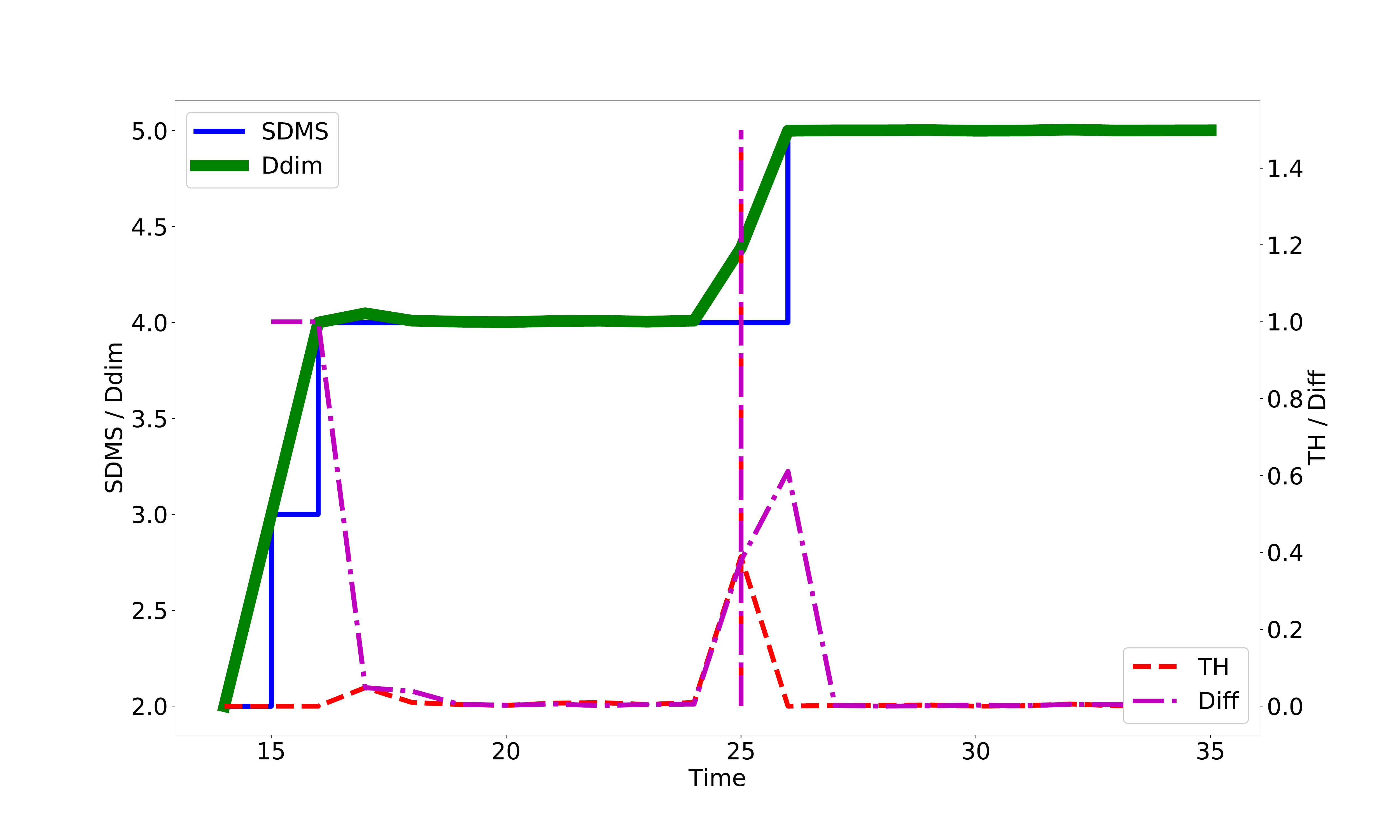}
          %./experiment20190123/ddimmarket/gmmcluddimrealdata-eps-converted-to.pdf}
          %gmm_clu_se_ddim_20180129-005731}
          %gmm_clu_se_ddim_real_data-eps-converted-to.pdf
      %\vspace*{-0.5cm}
          \caption{Change sign detection for market data}\label{market}
        \end{center}
        
      \end{figure}

\begin{table}[htb]
\begin{center}
\caption{Market structure change} \label{tb35}
%\hspace*{-1.5cm}
%\begin{minipage}{0.45\hsize}
\begin{center}
$t=24$ \\
\ \ \\
\begin{tabular}{|ccccc|}
% \multicolumn{2}{c}{} & 
\hline 
cat.& {\small c1}&{\small c2}  & {\small c3} & {\small c4}\\
\hline 
A& 0& 1& 1993&0 \\
B & 0& 2146& 0& 0 \\
C & 12& 25& 7& 156\\
D & 1768& 1& 7& 0 \\
\hline 
$\sharp$ &211 &126 & 138&437 \\
\hline
\end{tabular} \\
\ \ \\ \ \ \\
\end{center}
%\end{minipage}
%\begin{minipage}{0.45\hsize}
\begin{center}
$t=25$ \\
\ \ \\
\begin{tabular}{|ccccc|}
% \multicolumn{2}{c}{} & 
\hline 
cat.& {\small c1}&{\small c2}  & {\small c3} & {\small c4}\\
\hline 
A& 0& 1& 1959&0 \\
B & 0&2201 & 0&  0\\
C & 11& 21& 8& 184 \\
D & 1919& 1& 9& 0\\
\hline
$\sharp$ & 212&124 & 141& 435 \\
\hline
\end{tabular}\\ 
\ \ \\ \ \ \\
\end{center}
%\end{minipage} \\
%\begin{minipage}{0.8\hsize}
\begin{center}
$t=26$\\
\ \ \\
\begin{tabular}{|cccccc|}
% \multicolumn{2}{c}{} & 
\hline 
cat.& {\small c1}&{\small c2}  & {\small c3} & {\small c4}& {\small c5} \\
\hline 
A& 0& 0&2139 & 0& 0\\
B & 0& 2199& 1& 0& 0 \\
C & 12& 19& 8& 2962& 0\\
D &1916 &1& 12& 0 & 0\\
\hline 
$\sharp$ &213 &123 &148 & 283& 145\\
\hline
\end{tabular}\\ \ \\
\end{center}
%\end{minipage}
\end{center}
\end{table}

\subsubsection{Results}
Fig. \ref{market} shows Ddim~(green), estimated number of clusters in GMM (blue) using SDMS, and time points of alarms raised by 
TH and Diff~(red and purple) with $\delta _{1}=\delta _{2}=0.1$.
Table \ref{tb35} shows the clustering structures $t=24,25,26$. Each number in the $(i,j)$th cell shows the purchase volume of category $i\ (
=A,B,C,D)$ for the customers in  the $j$th cluster $cj\ (j=1,2,3,4)$. The last row shows the number of customers.

The purchase volume of category $C$ in cluster $c4$ gradually increased from $t=24$ to $t=25$, eventually $c4$ started to collapse at $t=25$ and was split into $c4$ and $c5$ at $t=26$.  
We confirm from Table \ref{tb35} that 
$c4$ consisted  of heavy users in category $C$,  at $t=26$, some of them became dormant users that did not purchase anything to form a new cluster. 
The SDMS algorithm detected this market structure change at $t=26$. 
As shown in Fig. \ref{market}, 
TH and Diff successfully raised an alarm at $t=25$ as a sign of that market structure change. 
The reason why we could detect the early warning signal is that there were gradual changes among clusters as well as within individual clusters before the clustering change occurred.
Our result shows that our method was effective in detecting signs of model changes for such a case.

 \subsection{Real Data:  Electric Power Consumption Data}

\subsubsection{Data sets}
Next we apply our method to the household electric power consumption dataset provided by \cite{power}. This
dataset contains three categories of electric power consumption corresponding to electricity consumed 1) in  kitchen and laundry rooms,  2) by electric water heaters and 3) by air-conditioners. The data 
were obtained every other minute from Dec. 17, 2006 to Dec. 10, 2010.  We
set ${\bm x}_{t} = (x_{1}, \cdots , x_{n})$ and 
$x_{i} = (x_{i1}, x_{i2}, x_{i3})$ where each $x_{i}$ denotes the value of consumption per an hour for the three categories, respectively and ${\bm x}_{t}$ is the value of  consumption for two weeks ($n=336$). 

\subsubsection{Results}
Fig.\ref{consumption} shows how Ddim (the green curve) and the number of clusters (the blue line) changed over time. 
Here each cluster shows a consumption pattern.
The red dotted line shows the alert positions for TH and Diff with $\delta _{1}=\delta _{2}=0.1$.
%Note that TH and Diff made alerts at the same time. 
Let us focus on the duration from $t=18$ to $t=22$. At $t=18,19$, there were three clusters, one of which collapsed to two clusters at $t=21$, eventually, produced the fourth cluster. The Ddim graph in Fig.\ref{consumption} shows that Ddim gradually increased from $k=3$ to $k=4$ during the period. 
The alert was made by TH and Diff at $t=20$ while there were still three clusters. This alert can be thought of as a sign of the emergence of a new cluster having a unique consumption pattern.  %These patterns show homogeneous consumption patterns with relatively larger weight for category 1 and 3, respectively.
% of the cluster collapse process and consumption patterns.

%Table \ref{tb2} shows the mean values and the data counts for individual clusters from $t=18$ (May 14th, 2007) to $t=23$(June 4th, 2007).
%\begin{comment}
\begin{figure}[!thb]
        \begin{center}
%        \hspace*{-5.0cm}
          %%% result for data set 1 for structural entropy
          \centering
          \includegraphics[keepaspectratio, 
          width=8.0cm] %, height=3.7cm] %height=3.0cm]
          {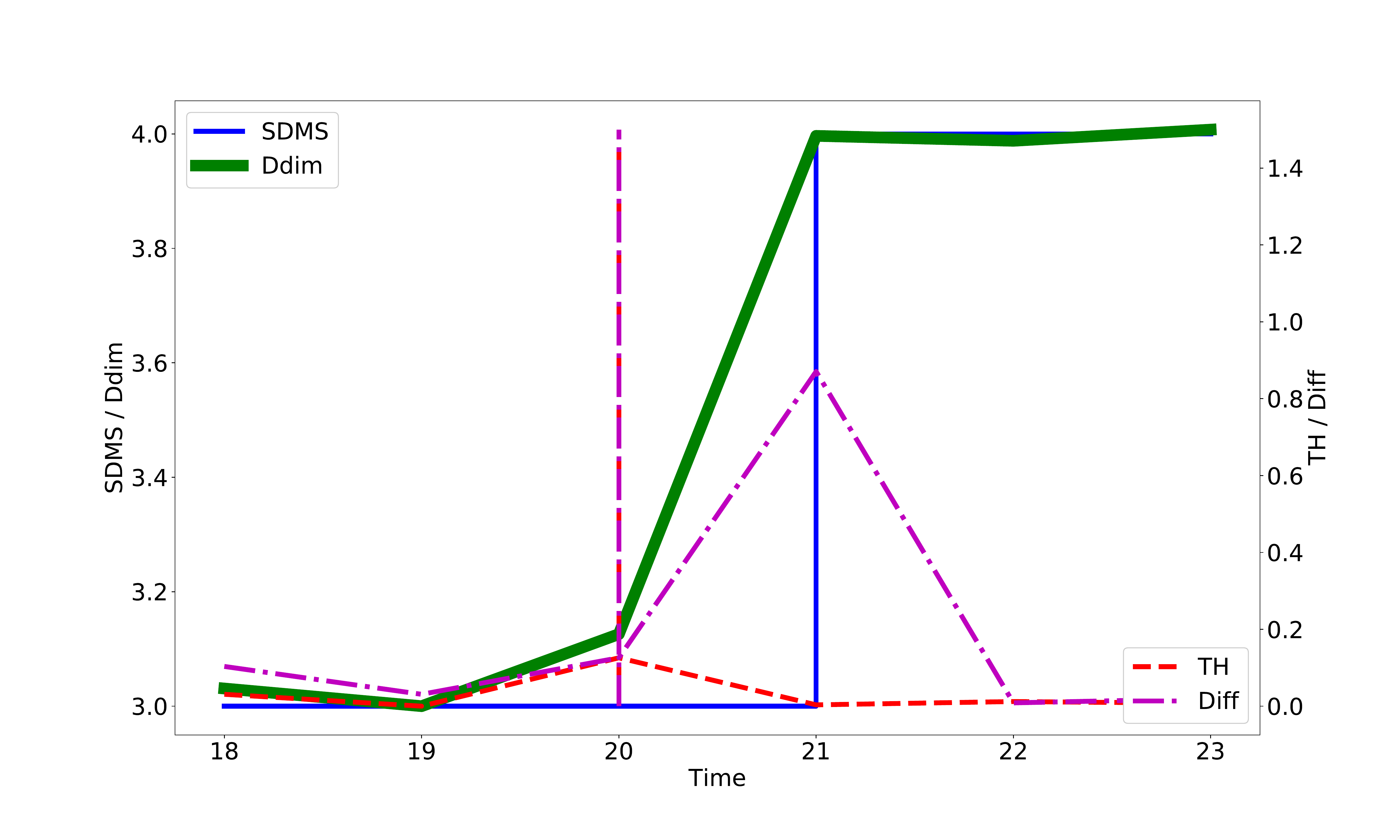}
          %./jpg/ddim/real_data/gmm_clu_ddim_realdata-eps-converted-to.pdf}   %gmm_clu_ddim_realdata_20181203-172724-eps-converted-to.pdf} 
%
          %gmm_clu_se_ddim_20180129-005731}
          %gmm_clu_se_ddim_real_data-eps-converted-to.pdf
          \vspace*{-0.5cm}
          \caption{Change sign detection for power consumption data}\label{consumption}
 %    \vspace*{-0.5cm}
        \end{center}
        
      \end{figure}
%\end{comment}
   \begin{table*}[!h]
      \caption{Electric power consumption structure change}
      \label{tab:household_result}
      \begin{minipage}{\hsize}
        %\subcaption{The week from May 14th in 2007}
        \label{tab:household_result_070514}
        \begin{center}
        The week from May 14th\\
          \begin{tabular}{|rrrrr|} \hline
            c & m1 & m2 & m3 & $\sharp$ \\ \hline 
            0 & 0.01 & 2.73 & 0.01 & 163 \\ 
            1 & 0.00 & 2.80 & 6.16 & 92 \\ 
            2 & 5.18 & 3.05 & 4.74 & 57 \\ \hline
          \end{tabular}\\
          \ \ \\
          \ \ \\
        \end{center}
      \end{minipage}
      \begin{minipage}{\hsize}
        %\subcaption{The week from May 21st(sign was detected by Ddim)}
        \label{tab:household_result_070521}
        \begin{center}
        The week from May 21st\\
          \begin{tabular}{|rrrrr|} \hline
            c & m1 & m2 & m3 & $\sharp$ \\ \hline 
            0 & 0.01 & 2.58 & 0.01 & 168 \\ 
            1 & 0.00 & 2.49 & 6.22 & 99 \\ 
            2 & 5.65 & 2.64 & 4.34 & 45 \\ \hline
          \end{tabular}\\
          \ \ \\
          \ \ \\
        \end{center}
      \end{minipage}
      \begin{minipage}{\hsize}
        %\subcaption{The week from May 28th(model change was detected by SDMS)}
        \label{tab:household_result_070528}
        \begin{center}
        The week from May 28th\\
          \begin{tabular}{|rrrrr|} \hline
            c & m1 & m2 & m3 & $\sharp$ \\ \hline 
            0 & 0.01 & 2.70& 0.01 & 150 \\ 
            1 & 0.00 & 3.01 & 6.38 & 99 \\ 
            2 & 1.73 & 2.72 & 5.87 & 8 \\ 
            3 & 6.20 & 3.11 & 4.91 & 55 \\ \hline
          \end{tabular}\
          \ \ \\
          \ \ \\
        \end{center}
      \end{minipage}
\begin{comment}
      \begin{minipage}{.48\hsize}
        \subcaption{2007年6月4日週}
        \label{tab:household_result_070604}
        \begin{center}
          \begin{tabular}{|r||r|r|r||r|} \hline
            c & meter\_1 & meter\_2 & meter\_3 & count \\ \hline \hline
            0 & 0.01 & 0.01 & 0.01 & 37 \\ \hline
            1 & 0.01 & 3.11 & 6.27 & 83 \\ \hline
            2 & 0.00 & 3.56 & 0.01 & 141 \\ \hline
            3 & 5.61 & 3.80 & 4.74 & 51 \\ \hline
          \end{tabular}
        \end{center}
      \end{minipage}
\end{minipage}
\end{comment}
    \end{table*}

Table \ref{tab:household_result} shows the contents of clusters on the weeks starting from May 14th, 21st, and 28th in 2007. $c$ means  clusters and $m1,m2,m3$ mean the mean amounts of meter 1,2,3, respectively. The last column shows the total amount of users in a respective cluster.
A sign of model change was detected on May 21st. The model change was detected on May 28th. We see from Table 
\ref{tab:household_result} that cluster 2 collapsed into clusters 2 and  3.
Cluster 2 shows a pattern of homogeneous consumption with a relatively high weight on category 3.  
Cluster 3 shows a pattern of homogeneous consumption with a relatively high weight on category 1.
The sign of this collapse was successfully detected on May 21st by monitoring the Ddim value.
The reason why we could detect the early warning signal is that there was a  gradual change in the collapse of cluster 2  before the clustering change occurred.
Our result shows that our method was effective in detecting signs of model changes for such a case.

\section{Relation of Ddim to MDL Learning}
 This section gives a theoretical foundation of Ddim by relating it to the rate of convergence of {the MDL learning algorithm~\cite{barron,yamanishi92}.} 
 It selects a model with the shortest total codelength required for encoding the data as well as the model itself.
We give an NML-based version of the MDL algorithm as follows.

Let ${\mathcal F}=\{{\mathcal P}_{1},\dots , {\mathcal P}_{s}\}$ where $\mid {\mathcal F}\mid =s< \infty$ and each ${\mathcal P}_{i}$ is a class of probability distributions. 
For a given training data sequence ${\bm x}=x_{1},\dots ,x_{n}$ where each $x_{i}$ is independently drawn, the MDL learning algorithm selects $\hat{{\mathcal P}}$ such that 
\begin{eqnarray}\label{mdllearning}
\hat{{\mathcal P}}&=&\argmin _{{\mathcal P}\in {\mathcal F}}(-\log p_{_{\rm NML}}({\bm x}; {\mathcal P}))\\
&=&
\argmin_{{\mathcal P}\in {\mathcal F}}\left\{-\log \max _{p\in {\mathcal P}}p({\bm x})+\log {\mathcal C}_{n}({\mathcal P})%+\lambda \ell({\mathcal P})
\right\}, \nonumber
%\Longrightarrow \min {\rm w.r.t.}\  {\mathcal P},
\end{eqnarray}
where ${\mathcal C}_{n}({\mathcal P})$ is the parametric complexity of ${\mathcal P}$ as in (\ref{int1}).
% and $\ell ({\mathcal P})$ is a codelength function satisfying the {\em Kraft's inequality}:
%\[\sum _{{\mathcal P}\in {\mathcal F}}e^{-\ell ({\mathcal P})}\leq 1.\]This is a necessary and sufficient condition for $\ell$ to define a uniquely decodable code \cite{cover}.
The MDL learning algorithm outputs the NML distribution  associated with $\hat{{\mathcal P}}$ as in (\ref{mdllearning}): For a sequence ${\bm y}=y_{1},\dots , y_{n}$ 
\begin{eqnarray}\label{mdloutput}
\hat{p}({\bm y})=\frac{\max _{p\in \hat{{\mathcal P}}}p({\bm y})}{C_{n}(\hat{{\mathcal P}})}.  
\end{eqnarray}
Note that ${\bm y}$ is independent of the training sequence ${\bm x}$ used to obtain $\hat{{\mathcal P}}$.
In previous work \cite{barron,yamanishi92}, the MDL learning algorithm has been designed so that it outputs the two-stage shortest codelength  distribution with quantized parameter values, belonging to the model classes.
Our algorithm differs from them in that it outputs the NML distribution (\ref{mdloutput}), which is not included in the model classes.
The NML distribution and the MDL principle are the central notions in deriving Ddim throughout this paper. Thus it is significant to investigate the relation of Ddim to the NML distribution estimated with the MDL learning algorithm.

 We have the following theorem relating Ddim to  the rate of convergence of the  MDL learning algorithm.
\begin{theorem}\label{rate2}
Suppose that each ${\bm x}$ is
generated according to %the true model
$ p^{*}\in {\mathcal P}^{*}\in {\mathcal F}=\{{\mathcal P}_{1},\dots , {\mathcal P}_{s}\}$.
%Let $\hat{{\mathcal P}}\in {\mathcal F}$ be the output of the MDL learning algorithm and
Let $\hat{p}$ be the output of the MDL learning algorithm
%the NML distribution associated with $\hat{{\mathcal P}}$
as in (\ref{mdloutput}).
Let $d_{B}^{(n)}(\hat{p},p^{*})$ be the Bhattacharyya distance
between $\hat{p}$ and $p^{*}$: \\
\begin{eqnarray}\label{bha}
d_{B}^{(n)}(\hat{p},p^{*})\buildrel \rm def \over =-\frac{1}{n}\log \sum _{{\bm y}} (p^{*}({\bm y})\hat{p}({\bm y}))^{\frac{1}{2}}.
\end{eqnarray}\\
Then for any $\epsilon >0$,
%$\epsilon >((1/2)\log C_{n}({\mathcal P}^{*})+ \log |{\mathcal F}|)/n$,
we have the following upper bound on
the probability that under the condition for ${\mathcal P}^{*}$ as in Theorem \ref{basic}, 
the Bhattacharyya distance between the output of the MDL learning algorithm and the true distribution exceeds $\epsilon$: 
\begin{eqnarray}\label{000}
Prob[d_{B}^{(n)}(\hat{p},p^{*})>\epsilon ]
%& <&  \exp \left(-n\epsilon +\frac{1}{2}\log C_{n}({\mathcal P}^{*})+\log \mid {\mathcal F} \mid \right)\nonumber \\
%\end{eqnarray}
%Further under the condition for ${\mathcal P}^{*}$ as in Theorem \ref{basic}, we have
%\begin{eqnarray}
\label{00111}
%Prob[d_{B}^{(n)}(\hat{p},p^{*})>\epsilon ]
& =&O\left(n^{{\rm Ddim}({\mathcal P}^{*})/4}e^{-n\epsilon}\right). 
\end{eqnarray}

Suppose that ${\mathcal P}$ is chosen randomly according to the probability distribution $\pi ({\mathcal P})$ over ${\mathcal F}=\{{\mathcal P}_1,\dots , {\mathcal P}_{s}\}$ and that the unknown
true distribution $p^{*}$ is chosen from ${\mathcal P}^{*}$.
Then we have the following upper bound on
the expected Bhattacharyya distance between the output of the MDL learning algorithm and the true distribution: 
\begin{equation}
E_{{\mathcal P}^{*}}E_{{\bm x}\sim p^{*}\in {\mathcal P}^{*}}[d_{B}^{(n)}(\hat{p},p^{*}) ]
\label{00001}
=O\left(\frac{{\rm Ddim}({\mathcal F}^{\odot})\log n}{n}\right),
\end{equation}
where ${\rm Ddim}({\mathcal F}^{\odot})$ is Ddim for model fusion as in (\ref{fusiondef}).
\end{theorem}
The proof is given in Appendix.  
This result may be generalized into the agnostic case where the model class misspecifies the true distribution (see also \cite{ding1} for this case)). We omit this result from this manuscript since our main concern is how the expected generalization performance is related to Ddim.

Theorem \ref{rate2} implies that %if $\epsilon >((1/2)\log C_{n}({\mathcal P}^{*})+\log |{\mathcal F}|)/n$, 
the NML distribution with model of the shortest NML codelength converges exponentially to the true distribution in probability as $n$ increases and the rate is governed by Ddim for the true model.
In conventional studies on PAC~(probably approximately correct) learning~\cite{Haussler}, the performance of the empirical risk minimization algorithm has been analyzed using the technique of {\em uniform convergence}, where the rate of convergence is governed by the metric dimension.
Meanwhile, the performance of the MDL learning algorithm is  analyzed using the {\em non-uniform convergence} technique,
since the non-uniform model complexity
is considered. In this case the rate of convergence of the MDL algorithm is governed by Ddim.
Then the expected Bhattacharyya distance between the true distribution and the output of the MDL learning algorithm is characterized by Ddim for model fusion over ${\mathcal F}$.

\section{Conclusion}
This paper has proposed a novel methodology for  detecting signs of model changes from a data stream. 
The key idea is to conduct continuous model selection using the notion of descriptive dimensionality~(Ddim). 
Ddim quantifies the real-valued model dimensionality in the model  transition period. 
We are able not only to visualize the model complexity in the transition period of model changes, but also to detect their signs by tracking the rise-up of Ddim.
Focusing on the model changes in Gaussian mixture models, we have shown that gradual structure changes of GMMs can be effectively visualized by drawing a Ddim graph.
Furthermore, we have empirically demonstrated that our methodology was able to detect signs of changes of the number of mixtures in GMM and those of the order of AR model earlier than  they were actualized. 
Experimental results have shown that it was able to detect them significantly earlier than any other existing dynamic model selection methods.

%Note that unlike the existing measures such as structural entropy as in \cite{bigdata2018}, etc., Ddim has a clear meaning of model dimensionality, which visualizes how rapidly the model changes over time.

%Although we focused on GMM for the knowledge representation, our methodology can be straightforwardly applied to other models of probability distributions. 
%Empirical validation of our methodology for other kinds of models has remained for future studies.
%Our methodology suggests some  possibility of changing the notion of model selection; from discrete model selection to continuous one.
This paper has offered the use of continuous model change selection in the scenario of model change sign detection only. 
Exploring other scenarios of continuous model selection has remained for future studies.

  \section*{Acknowledgment}

This work was partially supported by 
JST KAKENHI 191400000190
and JST-AIP JPMJCR19U4.

\begin{comment}
\ifCLASSOPTIONcompsoc
  \section*{Acknowledgments}
\else
  \section*{Acknowledgment}
\fi

This work was partially supported by 
JST KAKENHI 191400000190
and JST-AIP JPMJCR19U4.
\end{comment}

%\newpage

\appendix

\section{Proof of Theorem 1}

Let ${\mathcal P}$ be a $k$-dimensional parametric class, which we denote as ${\mathcal P}_k=\{ p({\bm x};\theta ):\  \theta \in \Theta _{k}\}$ where $\Theta _k$ is a $k$-dimensional parametric space. In this case, we denote $g(\hat{\theta}, \theta )$ instead of $g(\hat{p}, p)$.
Let the finite set of  $k$-dimensional real-valued parameters space be
$\overline{\Theta} _{k}=\{\theta _{1},\theta _{2},\dots\}$ and let
$\overline{{\mathcal P}}_{k}=\{ p({\bm x};\theta ):\  \theta \in \overline{\Theta} _{k}\subset {\mathbb R}^{k}\}.$
Let $I_{n}(\theta )$ be the Fisher information matrix at $\theta$: $I_{n}(\theta )\buildrel \rm def \over =(1/n)E_{\theta}[\partial ^{2}(-\log p({\bm x}; \theta ))/\partial \theta \partial \theta ^{\top}]$ and  suppose that $\lim _{n\rightarrow \infty}I_{n}(\theta )=I(\theta )$ for each $\theta$.
Below we denote $p({\bm x};\theta)$ as $p_{\theta}$.
Consider $\epsilon ^{2}$-neighborhood of $\theta _{i}$ with respect to the KL divergence $d_{n}$:
\begin{eqnarray*}D_{\epsilon}(i)\buildrel \rm def \over =\{\theta : d_{n}(p_{\theta _{i}}, p_{\theta})\leq \epsilon ^{2}\}.
\end{eqnarray*}
Note that $d_{n}(p_{\theta _{i}},p_{\theta})$ is written using Taylor's expansion up to the second order
as follows: Under the condition that $\log p$ is three-times differentiable, $\max _{a,b,c}\mid \partial ^{3}\log p({\bm x};\theta )/\partial \theta _{a}\partial \theta _{b}\partial \theta _{c}\mid <\infty$, 
\begin{eqnarray*}
d_{n}(p_{\theta _{i}},p_{\theta})
& =&%\frac{1}{n}E_{\theta _{i}}[\log p({\bm x};\theta _{i})]-\frac{1}{n}E_{\theta _{i}}[\log p({\bm x};\theta _{i})]
-\frac{1}{n}E_{\theta_{i}}\left[\frac{\partial \log p({\bm x};\theta)}{\partial \theta}\mid _{\theta_{i}}\right](\theta -\theta _{i}) \\
& & +\frac{1}{2n}(\theta -\theta _{i}) ^{\top}E_{\theta _{i}}\left[ -\frac{\partial ^{2}\log p({\bm x};\theta )}{\partial \theta \partial \theta ^{\top}}\mid _{\theta_{i}}\right](\theta -\theta _{i}) +O(\parallel \theta-\theta _{i}\parallel ^{3})\\
&& =\frac{1}{2}(\theta-\theta  _{i})^{\top}I_{n}(\theta _{i})(\theta -\theta _{i})+O(\parallel \theta-\theta_{i}\parallel ^{3}),
\end{eqnarray*}
where we have used the fact:
\begin{eqnarray*}
E_{\theta_{i}}\left[\frac{\partial \log p({\bm x};\theta)}{\partial \theta}\mid _{\theta_{i}}\right]&=&\sum _{\bm x}p({\bm x};\theta _{i})\frac{\partial \log p({\bm x};\theta )}{\partial \theta }\mid _{\theta_{i}}\\
&=&\frac{\partial \sum _{\bm x}p({\bm x};\theta )}{\partial \theta} \mid _{\theta_{i}}\\ &=&0.
\end{eqnarray*}
Therefore, we may consider $\tilde{D}_{\epsilon}(i)$ in place of $D_{\epsilon}(i)$.
\begin{eqnarray*}
\tilde{D}_{\epsilon}(i)=\{\theta : (\theta -\theta _{i})^{\top}I_{n}(\theta _{i})(\theta -\theta _{i})\leq C\epsilon ^{2}\},
\end{eqnarray*}
where $C$ does not depend on $n$ nor $\epsilon$.
Let $B_{\epsilon}(i)$ be the largest hyper-rectangle within $\tilde{D}_{\epsilon}(i)$ centered at $\theta _{i}$.
For some $1\leq  C'< \infty$, for any $i$, we have 
\begin{eqnarray}\label{volume}
\mid B_{\epsilon}(i)\mid \leq \mid \tilde{D}_{\epsilon}(i)\mid \leq C'\mid B_{\epsilon}(i)\mid .
\end{eqnarray}

Along with  \cite{rissanen} (p.74), geometric analysis of $B_{\epsilon}(i)$ yields the Lebesgue volume of $B_{\epsilon}(i)$ as follows:
\begin{eqnarray*}
\mid B_{\epsilon}(i)\mid &=&\left( \frac{4C\epsilon ^{2}}{k}\right)^{\frac{k}{2}}\mid I_{n}(\theta _{i})\mid ^{-\frac{1}{2}}=2^{k}\prod ^{k}_{j=1}\sqrt{\frac{C\epsilon ^{2}}{k\lambda_ {j}}},
\end{eqnarray*}
where $\lambda _{j}$ is the $j$-th largest eigenvalue of $I_{n}(\theta _{i})$.

We choose $\overline{\Theta }_{k}$ so that the central limit theorem holds. Then
%in the form of (\ref{clt}),  
for sufficiently large $n$, as $\theta \rightarrow \theta _{i}$,
\begin{eqnarray*}
g(\theta _{i}, \theta )\simeq \left( \frac{n}{2\pi}\right)^{\frac{k}{2}} \mid I(\theta _{i})\mid ^{\frac{1}{2}}e^{-n(\theta-\theta _{i})^{\top}I_{n}(\theta _{i})(\theta-\theta _{i})}.
\end{eqnarray*}
%This choice is possible since $\theta _{i}$ is one of the maximum likelihood estimate of $\theta$.
Thus for $\theta \in D_{i}(\epsilon)$, we obtain
\begin{eqnarray}\label{gfunc25}
%g(\theta _{i},\theta_{i})|B_{\epsilon}(i)|\simeq \left(
%\frac{2C\epsilon ^{2}n}{k\pi}\right)^{\frac{k}{2}}.
g(\theta ,\theta )\mid B_{\epsilon}(i)\mid \simeq \left(
\frac{2C\epsilon ^{2}n}{k\pi}\right)^{\frac{k}{2}}.
\end{eqnarray}

Next define $Q_{\epsilon}(i)$ as
\begin{eqnarray*}
Q_{\epsilon}(i)\buildrel \rm def \over =\int _{\hat{\theta}\in \tilde{D}_{\epsilon}(i)}g(\hat{\theta}, \hat{\theta})d\hat{\theta}.
\end{eqnarray*}
and let
$m_{n}(\epsilon )$ be the smallest number of elements in $\overline{{\Theta}}_{k}$:
\begin{eqnarray}\label{vv}
\log C_{n}(k)\leq
\log \sum ^{m_{n}(\epsilon )}_{i=1}Q_{\epsilon}(i).
\end{eqnarray}
Combining (\ref{gfunc25}) and (\ref{volume}) with (\ref{vv}) yields
\begin{eqnarray}
\log C_{n}({\mathcal P}_k) 
&=&\log m_{n}(\epsilon )+\sup _{\overline{\Theta }_{k}}\left\{\frac{k}{2}\log \left( \frac{2C\epsilon ^{2}n}{k\pi}\right)\right\}+O(1) \label{o1} \\
&=&\log m_{n}(\epsilon )+\frac{k}{2}\log ({\epsilon ^{2}n}) +O(1).\label{o2}
\end{eqnarray}
where $C$ in (\ref{o1}) depends on $\overline{\Theta}_{k}$  and the $O(1)$ term in (\ref{o2}) may depend on $k$, but both of them do not depend on $n$ nor $\epsilon$.
The supremum in (\ref{o1}) is taken with respect to $\bar{\Theta}_{k}$ so that (\ref{gfunc25}) holds.
Setting $\epsilon ^{2}n=O(1)$ yields 
\begin{eqnarray*}\label{nmb1}
\log C_{n}({\mathcal P}_{k}) &= &\log m_{n}(\epsilon :{\mathcal P}_{k})+\frac{k}{2}\log (\epsilon ^{2}n)+O(1)\\
&=&\log m_{n}(1/\sqrt{n} :{\mathcal P}_{k})+O(1).
%-\frac{k}{2}\log \left( \frac{2\epsilon ^{2}n}{k\pi}\right)+
\end{eqnarray*}
This completes the proof of Theorem 1.
\hspace*{\fill}$\Box$

\section{Proof of Theorem 3}
Let $p^{*}$ be the true distribution %$p(x^{n}:{\mathcal P}^{*})$
 associated with the true model ${\mathcal P}^{*}$.
Let $\hat{{\mathcal P}}$ be the model selected by the MDL learning algorithm and let
$p_{_{\rm NML}}({\bm x};\hat{{\mathcal P})}$ be the NML distribution associated with $\hat{{\mathcal P}}$. We write it as $\hat{p}$.
We employ the proof technique similar to that for two-part code estimators in \cite{barron,yamanishi92}.

By the definition of the MDL learning algorithm,
we have
\begin{eqnarray}\label{in}
\min_{{\mathcal P}}( -\log p_{_{\rm NML}}({\bm x}; {\mathcal P}))
& \leq &-\log p_{_{\rm NML}}({\bm x}; {\mathcal P}^{*})\nonumber \\
& =&-\log \max _{p\in {\mathcal P}^{*}}p({\bm x})+\log C_{n}({\mathcal P}^{*})\nonumber \\
& \leq &
-\log p^{*}({\bm x})+\log C_{n}({\mathcal P}^{*}). 
\end{eqnarray}

Let $p_{_{\rm NML},{\mathcal P}}$ be the NML distribution $p_{_{\rm NML}}({\bm x}:{\mathcal P})$ associated with ${\mathcal P}.$ %defined as
%\begin{equation}
%p_{_{\rm NML}}({\bm x}:{\mathcal P})=\frac{\max _{p\in {\mathcal P}}p({\bm x})}{C_{n}({\mathcal P})}. \label{nmld}
%\end{equation}
For $\epsilon >0$, the following inequalities hold:
\begin{align}%\label{event}
&Prob[d_{B}^{(n)}(\hat{p},p^{*})>\epsilon ] \nonumber \\
&\leq
Prob[ {\bm x}: \ (\ref{in})\ {\rm holds\ under}\ d_{B}^{(n)}(\hat{p},p^{*})>\epsilon ] \nonumber \\
&= Prob\Biggl[ {\bm x}: \min _{{\mathcal P}:d_{B}^{n}(p_{_{\rm NML},{\mathcal P}},p^{*})>\epsilon }(-\log p_{_{\rm NML}}({\bm x};{\mathcal P})) \nonumber \\
&\ \ \ \ \ \ \ \ \ \ \ \ \ \ \ \ \ \ \leq -\log p^{*}({\bm x})+\log C_{n}({\mathcal P}^{*})\Biggr] \nonumber \\
&=Prob\Biggl[ {\bm x}: \max _{{\mathcal P}:d_{B}^{n}(p_{_{\rm NML},{\mathcal P}},p^{*})>\epsilon }p_{_{\rm NML}}({\bm x}:{\mathcal P})\geq \frac{p^{*}({\bm x})}{C_{n}({\mathcal P}^{*})}\Biggr]\nonumber \\
&\leq \sum _{{\mathcal P}\in {\mathcal F}, d_{B}^{(n)}(p_{_{\rm NML},{\mathcal P}}, p^{*})>\epsilon}Prob\Biggl[ {\bm x}:
%-\log
%p_{_{\rm NML}}({\bm x};{\mathcal P})\nonumber \\ \label{event00}
%&\ \ \ \ \ \ \ \ \ \ \ \ \ \ \ \ \ \ \ \ \ \ \ \ \ \ \ \
%\leq -\log p^{*}({\bm x})+
%\log C_{n}({\mathcal P}^{*}) \bigr],\\
p_{_{\rm NML}}({\bm x}:{\mathcal P})\geq\frac{ p^{*}({\bm x})}{C_{n}({\mathcal P}^{*})} \Biggr]. \label{sumevent00}%\label{event}
\end{align}

Let $E_{n}({\mathcal P})$ be the event:
%\begin{eqnarray*}
$p_{_{\rm NML}}({\bm x}:{\mathcal P})
\geq p^{*}({\bm x})/C_{n}({\mathcal P}^{*}).$
%\end{eqnarray*}
Under $E_{n}({\mathcal P})$,
\[1\leq \left(\frac{p_{_{\rm NML}}({\bm x};{\mathcal P})}{p^{*}({\bm x})}\right)^{\frac{1}{2}} ( C_{n}({\mathcal P}^{*}))^{\frac{1}{2}}. \]
Then under the condition that $d_{B}^{(n)}(p_{_{\rm NML},{\mathcal P}}, p^{*})>\epsilon$, we have
\begin{eqnarray}
Prob[E_{n}({\mathcal P})]&=&\sum _{{\bm x}\cdots E_{n}({\mathcal P})}p^{*}({\bm x})\nonumber \\
&\leq &\sum _{{\bm x}\cdots E_{n}({\mathcal P})}p^{*}({\bm x})\left(\frac{p_{_{\rm NML}}({\bm x}; {\mathcal P})}{p^{*}({\bm x})}\right)^{\frac{1}{2}} ( C_{n}({\mathcal P}^{*}))^{\frac{1}{2}} \nonumber \\
&\leq& \left\{\sum_{{\bm y}} (p_{_{\rm NML}}({\bm y};{\mathcal P})p^{*}({\bm y}))^{\frac{1}{2}} \right\} (C_{n}({\mathcal P}^{*}))^{\frac{1}{2}} \nonumber \\
&< &\exp (-n\epsilon+(\log C_{n}({\mathcal P}^{*}))/2) ,
\label{event100}
\end{eqnarray}
where  we have used the fact that under $d_{B}^{(n)}(p_{_{\rm NML},{\mathcal P}},p^{*})>\epsilon$, it holds 
\[\sum _{{\bm y}} (p_{_{\rm NML}}({\bm y};{\mathcal P})p^{*}({\bm y}))^{\frac{1}{2}}< e^{-n\epsilon }.\]
Plugging (\ref{event100}) into (\ref{sumevent00}) yields
\begin{eqnarray}
Prob[d_{B}^{(n)}(\hat{p},p^{*})>\epsilon ] 
&\leq &\sum _{{\mathcal P}\in {\mathcal F}, d_{B}^{(n)}(p_{_{\rm NML},{\mathcal P}}, p^{*})>\epsilon}Prob[E_{n}({\mathcal P})]\nonumber \\
&< &\sum _{{\mathcal P}\in {\mathcal F}, d_{B}^{(n)}(p_{_{\rm NML},{\mathcal P}}, p^{*})>\epsilon}\exp \left(-n\epsilon +(1/2)\log C_{n}({\mathcal P}^{*})\right)\nonumber \\
&\leq& \sum _{{\mathcal P}\in {\mathcal F}} \exp \left(-n\epsilon +(1/2)\log C_{n}({\mathcal P}^{*})\right) \nonumber \\
&=&
\exp \left(-n\epsilon +(1/2)\log C_{n}({\mathcal P}^{*})+\log \mid {\mathcal F}\mid \right). \label{por}
\end{eqnarray}
%where we used the Kraft's inequality: For $\lambda >2$,\\
%\begin{eqnarray*}
%\sum _{{\mathcal P}}\exp (-\lambda \ell ({\mathcal P})/2)\leq 1.
%\end{eqnarray*}
Under the condition for ${\mathcal P}^{*}$ as in Theorem 3, we have
\begin{eqnarray}\label{mk}
\frac{1}{2}\log C_{n}({\mathcal P}^{*})+\log \mid {\mathcal F}\mid =\frac{1}{4}{\rm Ddim}({\mathcal P}^{*})\log n+o(\log n).
\end{eqnarray}
Plugging (\ref{mk}) into (\ref{por})  yields (\ref{00111}).

%Note that by  (\ref{defdim2}) and
Let $r_{n}({\mathcal P}^{*})\buildrel \rm def \over =\{(1/2)\log {\mathcal C}_{n}({\mathcal P}^{*})+\log \mid {\mathcal F}\mid \}/n$.
For fixed ${\mathcal P}^{*}$, we have the following upper bound on the expected Bhattacharyya distance:
 \begin{eqnarray}
E_{{\bm x}\sim p^{*}\in {\mathcal P}^{*}}[d_{B}^{(n)}(\hat{p},p^{*})-r_{n}({\mathcal P}^{*})]
&=&\int ^{\infty}_{0}Prob[d_{B}^{(n)}(\hat{p},p^{*})-r_{n}({\mathcal P}^{*})>\epsilon ]d\epsilon \nonumber \\
%&=\int ^{\infty}_{0}Prob[d_{B}^{(n)}(\hat{p},p^{*})>\epsilon +r_{n}({\mathcal P}^{*})]d\epsilon \nonumber \\
&\leq &\int ^{\infty}_{0}e^{-n\epsilon}d\epsilon =\frac{1}{n},\label{km}
\end{eqnarray}
where we have used (\ref{00111}) to derive (\ref{km}).
Therefore, we have
\begin{eqnarray*}
E_{{\bm x}\sim p^{*}\in {\mathcal P}^{*}}[d_{B}^{(n)}(\hat{p},p^{*})]&\leq&r_{n}({\mathcal P}^{*})+\frac{1}{n}. 
%O\left( \frac{\log C_{n}({\mathcal P}^{*})}{n}\right),\\
%&=&O\left( \frac{{\rm Ddim}({\mathcal P}^{*})\log n}{n}\right).\\
 \end{eqnarray*}
Taking the expectation with respect to ${\mathcal P}^{*}$ yields
% where we used the fact that $L_{n}({\mathcal P}^{*})=O({\rm Ddim}({\mathcal P}^{*})\log n)$.
\begin{eqnarray}
E_{{\mathcal P}^{*}}[d_{B}^{(n)}(\hat{p},p^{*})]
&\leq &E_{{\mathcal P}^{*}}E_{{\bm x}\sim p^{*}\in {\mathcal P}^{*}}[r_{n}({\mathcal P}^{*})]+\frac{1}{n} \nonumber \\
%& =& \frac{E_{{\mathcal P}^{*}}[\log C_{n}({\mathcal P}^{*})]}{2n}+\frac{\log \mid {\mathcal F}\mid +1}{n} \nonumber \\
%&\leq & \frac{\log E_{{\mathcal P}^{*}}[C_{n}({\mathcal P}^{*})]}{2n} +\frac{\log |{\mathcal F}|+1}{n} \label{eq1}\\
&=&O\left(\frac{ E_{{\mathcal P}^{*}}[\log m_{n}(1/\sqrt{n},{\mathcal P}^{*})]}{n}\right) \label{eq2} \nonumber \\
&=&O\left(\frac{{\rm Ddim} ({\mathcal F}^{\odot})\log n}{n}\right).
\label{eq3}
 \end{eqnarray}
% Under the assumption as in Theorem \ref{rate2},
%\begin{eqnarray*}\\
%{\rm Ddim}({\mathcal P}^{*})= O\left(\frac{\log C_{n}({\mathcal P}^{*})}{\log n}\right).
%\end{eqnarray*}\\
%We have used the Jensen's inequality to derive (\ref{eq1}).
To derive (\ref{eq2}), we have used the fact $\log C_{n}({\mathcal P}^{*})=\log m_{n}(1/\sqrt{n},{\mathcal P}^{*})+O(1)$.
 This completes the proof.
\hspace*{\fill}$\Box$

\end{document}